\PassOptionsToPackage{table}{xcolor}
\documentclass[10pt,journal,compsoc]{IEEEtran}
\ifCLASSOPTIONcompsoc
  \usepackage[nocompress]{cite}
\else
  \usepackage{cite}
\fi
\ifCLASSINFOpdf
\else
\fi
\usepackage{epsfig}
\usepackage{graphicx}
\usepackage{amsmath}
\usepackage{amssymb}
\usepackage{comment}
\usepackage{multirow}
\usepackage{booktabs}
\usepackage{array, caption, threeparttable}
\usepackage{caption}
\usepackage{algorithm}
\usepackage{listings}
\usepackage{color}
\usepackage{mathtools}
\usepackage{todonotes}
\usepackage{gensymb}
\usepackage{url}

\def\ie{{\em i.e.}}
\def\eg{{\em e.g.}}
\def\etal{{\em et al.}}
\def\vs{\emph{vs.}}

\begin{document}
\title{Benchmarking Unified Face Attack Detection via Hierarchical Prompt Tuning}  
\author{
		Ajian Liu, 
		Haocheng Yuan, 
		Xiao Guo, 
		Hui Ma, 
		Wanyi Zhuang, 
		Changtao Miao, 
		Yan Hong, 
		Chuanbiao Song, 
		Jun Lan, 
		Qi Chu, 
		Tao Gong, 
		Yanyan Liang, 
		Weiqiang Wang, 
		Jun Wan,~\IEEEmembership{Senior Member,~IEEE,} \\
		Jiankang Deng,~\IEEEmembership{Senior Member,~IEEE,}
		Xiaoming Liu,~\IEEEmembership{Fellow,~IEEE,}
		Zhen Lei,~\IEEEmembership{Fellow,~IEEE}
	\IEEEcompsocitemizethanks{
	\IEEEcompsocthanksitem Ajian Liu and Haocheng Yuan are co-first authors. The corresponding author is Jun wan.
	\IEEEcompsocthanksitem Ajian Liu, Jun Wan and Zhen Lei are with the State Key Laboratory of Multimodal Artificial Intelligence Systems (MAIS), Institute of Automation, Chinese Academy of Sciences (CASIA), Beijing 100190, China (e-mail: \{ajian.liu, jun.wan, zhen.lei\}@ia.ac.cn).
	\IEEEcompsocthanksitem Haocheng Yuan is with the Department of Computer Science, College of Computing, City University of Hong Kong, Hong Kong, China.
	\IEEEcompsocthanksitem Hui Ma and Yanyan Liang are with the School of Computer Science and Engineering, Faculty of Innovation Engineering, Macau University of Science and Technology, Macau, China.
	\IEEEcompsocthanksitem Xiao Guo and Xiaoming Liu are with the Department of Computer Science and Engineering, Michigan State University, USA.
	\IEEEcompsocthanksitem Wanyi Zhuang, Changtao Miao, Qi Chu and Tao Gong are with the School of Cyber Science and Technology, University of Science and Technology of China, Hefei 230022, China.
	\IEEEcompsocthanksitem Yan Hong, Chuanbiao Song, Jun Lan and Weiqiang Wang are with the Ant Group, Zhejiang 310024, China.
	\IEEEcompsocthanksitem Jiankang Deng is with Imperial College London, London, UK.
}
}
\markboth{Journal of \LaTeX\ Class Files,~Vol.~*, No.~*, 2015}%
{Shell \MakeLowercase{\textit{et al.}}: Bare Advanced Demo of IEEEtran.cls for IEEE Computer Society Journals}

\IEEEtitleabstractindextext{%
\begin{abstract}
Presentation Attack Detection (PAD) and Face Forgery Detection (FFD) are proposed to protect face data from physical media-based Presentation Attacks (PAs) and digital editing-based DeepFakes (DFs), respectively. However, isolated training of these two models significantly increases vulnerability towards unknown attacks, burdening deployment environments. The lack of a Unified Face Attack Detection (UAD) model to simultaneously handle attacks in these two categories is mainly attributed to two factors: (1) A benchmark that is sufficient for models to explore is lacking. Existing UAD datasets only contain limited attack types and samples, leading to the model's confined ability to address abundant advanced threats. In light of these, through an explainable hierarchical way, we propose the most extensive and sophisticated collection of forgery techniques available to date, namely UniAttackDataPlus (\textbf{UniAttackData+}). Our UniAttackData+ encompasses $2,875$ identities and their $54$ kinds of corresponding falsified samples, in a total of $697,347$ videos. (2) The absence of a trustworthy classification criterion. Current methods endeavor to explore an arbitrary criterion within the same semantic space, which fails to exist when encountering diverse attacks. Thus, we present a novel Visual-Language Model-based Hierarchical Prompt Tuning Framework (\textbf{HiPTune}) that adaptively explores multiple classification criteria from different semantic spaces. Specifically, we construct a Visual Prompt Tree (\textbf{VP-Tree}) to explore various classification rules hierarchically. Then, by adaptively pruning the prompts, the model can select the most suitable prompts guiding the encoder to extract discriminative features at different levels in a coarse-to-fine manner. Finally, to help the model understand the classification criteria in visual space, we propose a Dynamically Prompt Integration (\textbf{DPI}) module to project the visual prompts to the text encoder to help obtain a more accurate semantics. Extensive experiments on 12 datasets have verified the potential to inspire further innovations in the UAD field. The UniAttackDataPlus dataset can be found at https://sites.google.com/view/face-anti-spoofing-challenge/dataset-download/uniattackdata-iccv2025, and the codes can be found at https://github.com/HaochengYUAN-Allen/UniAttackDataPlus-HiPTune.
\end{abstract}

}
\maketitle
\begin{figure*}[t]
	\centering
	\includegraphics[width=1.00\linewidth]{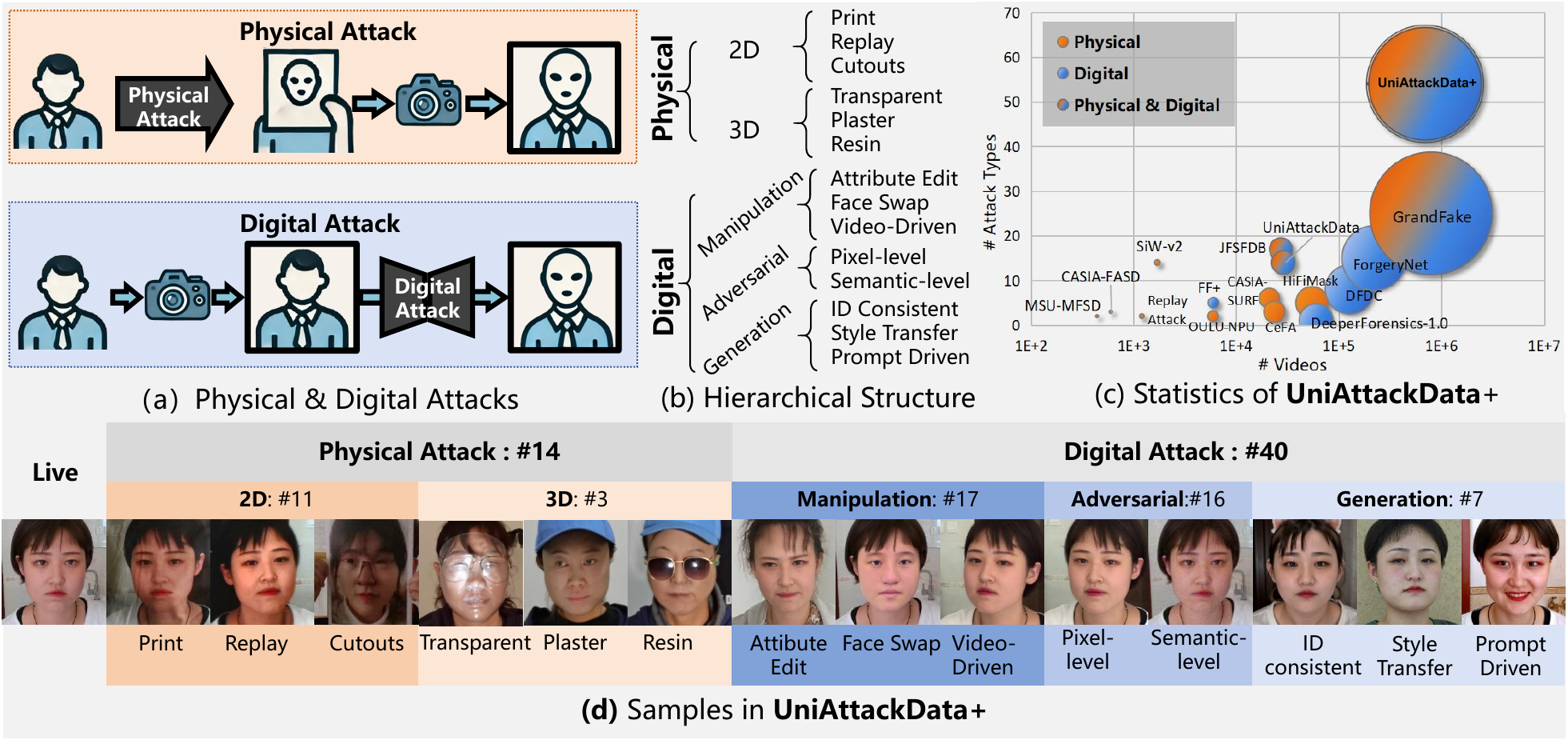}
	\caption{
		(a) The Physical and Digital Attack construction approaches. Physical Presentation Attacks aim to steal identity information before the capture devices, and Digital Attacks attempt to edit the existing image to falsify the face data. 
		(b) Our UniAttackData+ is constructed in an explainable hierarchical way. The coarse-to-fine categories provide researchers with prior knowledge, which is helpful for the model designs. 
		(c) A comparison of HiFAB with existing FAS, FFD, and UAD datasets shows that HiFAB surpasses all others in terms of attack types and video quantity. The radius of the circle represents the scale of the dataset, which is the number of videos contained in the dataset, corresponding to the horizontal axis of the figure. (Note: GrandFake is image-based) The vertical axis represents the number of attack types included in the dataset. 
		(d) Sample demonstrations of the UniAttackData+. The samples are selected from the second level of the structure. The number following each category name indicates the number of attack types included.
	}
	\label{fig_data}
\end{figure*}

\IEEEdisplaynontitleabstractindextext
%
\IEEEpeerreviewmaketitle
\ifCLASSOPTIONcompsoc
\IEEEraisesectionheading{\section{Introduction}\label{sec:introduction}}
\else

\section{Introduction} \label{sec:introduction} \fi
\IEEEPARstart{F}{ace} data are widely used in fields such as facial payment, mobile unlocking, and access authentication, due to the discriminative and easily obtainable facial biometric features. However, attackers can forge or steal identities in two ways to engage in illicit activities. As shown in the upper left of Fig.~\ref{fig_data} (a), attackers can deceive face recognition systems by using physical media to impersonate others before information collection, a method known as physical attacks or presentation attacks. Additionally, attackers may employ digital forgery (commonly referred to as deepfake) to edit or manipulate collected images for fraudulent purposes. To address the aforementioned threats, Face Anti-Spoofing, also known as Presentation Attack Detection (PAD), and Face Forgery Detection (FFD) have been proposed to ensure the security of facial data. Essentially, both methods are formulated as binary classification tasks, where the model determines the authenticity of an image by outputting a probability indicating whether it is real or fake.

Researchers have proposed a wide variety of attack methods and have constructed multiple databases to advance the development of defense algorithms. Common physical presentation attacks are generally categorized into two types: 2D attacks (\ie, print attack~\cite{zhang2020casia} and video attack~\cite{Chingovska_BIOSIG-2012}), and 3D attacks (\ie, mask~\cite{liu2022contrastive} and fake head~\cite{george2019biometric}). Most models~\cite{Liu2018Learning,george2019deep,zhang2020face,yu2020nasfas,cai2020drl,liu2022disentangling,Wang_2022_CVPR} achieve authenticity classification by analyzing image texture details, focusing on inconsistent edges, color distortion, screen moiré patterns, and mask pouring traces. However, their generalization severely deteriorates when faced with unknown domain datasets. Currently, domain generalization based PAD algorithms~\cite{} have become a hot research topic, such as UDG-FAS~\cite{liu2023towards}, SA-FAS~\cite{sun2023rethinking}, IADG~\cite{zhou2023instance}, FLIP-MCL~\cite{srivatsan2023flip}, GAC-FAS~\cite{le2024grad}, and TTDG-V~\cite{zhou2024test}. 

Digital attack methods can be broadly classified into three categories: editing, adversarial attacks, and generative attacks. Digital editing attacks aim to modify image attributes~\cite{wang2021safa,choi2018stargan} (\eg, changing facial expressions, age) or perform face swapping~\cite{rosberg2023facedancer,chen2020simswap} (\eg, replacing the current face with another), video-driven manipulation~\cite{hong2022depth,wang2021one} (\eg, using a video to drive facial expressions). Adversarial attacks can be divided into pixel-level~\cite{zou2022making,yan2022ila,lin2024boosting,schwinn2023exploring} and semantic-level attacks~\cite{luo2022frequency,wei2022towards,zhang2023transferable}, which either modify or erase relevant pixels or alter high-dimensional semantics to manipulate the original facial data. Digital generative attacks~\cite{huang2024consistentid,guo2024pulid,wang2024instantid}, the most recent and innovative type, involve generating entirely new facial data using generative models. To combat digital attacks, researchers~\cite{tian2024real,dong2023implicit,choi2024exploiting} typically focus on detecting inconsistencies in pixel-level and semantic-level information to achieve robust classification performance.
\begin{figure}[t]
	\centering
	\includegraphics[width=1\linewidth]{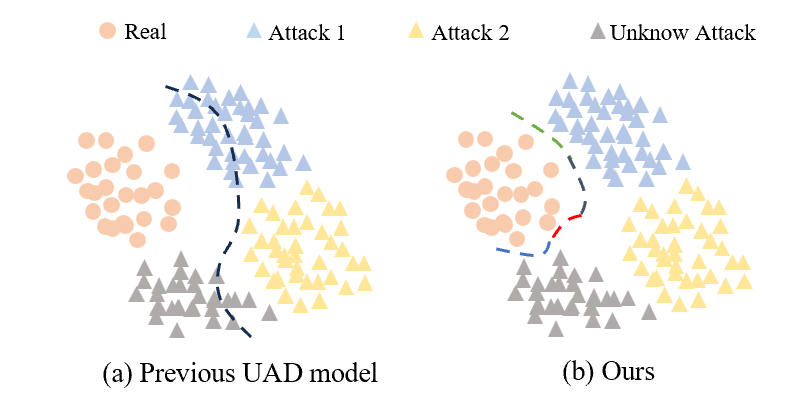}
	\caption{The dashed lines represent the classification criteria that separate real and fake faces in the feature space. (a) Existing models are trained to draw a single "line" in the feature space, leading to unpromising classification results. (b) Our HiPTune can adaptively integrate the classification criteria from different semantic spaces to achieve robust detection towards various attacks. Different colors represent distinct classification criteria.}
	\label{fig_Motivation}
\end{figure}

Existing PAD and FFD models are usually trained and deployed independently, leading to weak robustness against unknown or hybrid attacks and increased deployment complexity. A unified model that can leverage the advantages of different attack types while maintaining a lightweight parameter size is urgently needed. We identify two main reasons for the absence of such a solution: 

\textbf{(1) The lack of a comprehensive dataset that provides the model with a broader learning space.} Although there are currently three unified face attack detection datasets (\ie, GrandFake~\cite{deb2023unified} and JFSFDB~\cite{yu2024benchmarking}), they simply mix existing datasets that are tailored to specific attack types. This approach introduces substantial noise, such as variations in style, background, and identity, making it difficult to accurately evaluate the model's superiority and generalization capability. Furthermore, as shown in Fig.~\ref{fig_data} (b), their sizes are relatively small, containing limited attack types, and many of them are not publicly accessible, making them difficult to utilize effectively. To address these challenges, we previously introduced UniAttackData~\cite{fang2024unified}, a unified dataset featuring 14 attack types, including 2 physical and 12 digital attacks, all applied to the same identity. This design ensures identity consistency across diverse attacks, providing a valuable benchmark for training and evaluating generalized UAD models. However, its size and the diversity of attack types remain insufficient, especially lacking the most advanced generative face attack methods~\cite{huang2024consistentid,guo2024pulid,wang2024instantid} currently available.

\textbf{(2) The absence of a reliable classification criterion that can accurately categorize each sample.} Existing methods typically aim to identify a single classification criterion within a single semantic space, as illustrated in Fig.~\ref{fig_Motivation} (a), where the model is trained to discover a single `line' that separates real and fake faces. However, in the context of UAD, this single classification criterion cannot robustly classify all sample types. From the perspective of data distribution, two major challenges arise: First, the diversity of attack types is constantly expanding, and these attacks increasingly resemble real samples across various dimensions. A single classification criterion is inadequate to address the rapidly evolving and sophisticated forgery techniques. Second, the significant feature differences between physical and digital attacks result in large intra-class variance among forged samples. The feature extractor struggles to effectively cluster them within a common space, making it difficult for a single classification criterion to determine the class of each sample accurately.

To address the dataset limitations mentioned above, we adopted the identity-consistent approach of UniAttackData~\cite{fang2024unified} and developed UniAttackDataPlus (\textbf{UniAttackData+}), which is currently the largest publicly available unified face attack detection dataset. It features the most extensive variety of attack types, minimal interference, and highly structured characteristics, making it a robust and comprehensive resource for training and evaluating face attack detection models. We collected data from $2,875$ participants representing three distinct ethnic groups (\ie, African, East Asian, and Central Asian), capturing $18,250$ authentic videos under various lighting conditions, backgrounds, and acquisition devices. For each participant, we applied $54$ different attack methods, including $14$ physical attacks and $40$ digital attacks, resulting in a total of $679,097$ forged videos, which is $32$ times more than the existing dataset. The statistical comparison is shown in Fig.~\ref{fig_data} (c). From each video, we randomly extracted at least $30$ frames to construct the dataset. The diverse acquisition backgrounds of authentic faces provide a more varied dataset, encouraging models to focus on discriminative features rather than irrelevant factors. Moreover, the extensive volume of data and the wide variety of attack types offer a more comprehensive and expansive learning space, motivating researchers to design more sophisticated and advanced models. Most importantly, UniAttackData+ applies a full range of attack methods to each unique identity, covering almost all known attack types. This not only prevents models from achieving high performance through `identity memorization', but also enhances the dataset’s applicability in real-world deployment scenarios. Additionally, our dataset is structured in a hierarchical manner, with all attack types categorized from coarse to fine, as depicted in Fig.~\ref{fig_data} (d). This hierarchical structure provides researchers with a more interpretable framework for designing corresponding algorithms. 

To address the issue of insufficient robustness caused by a single classification criterion in UAD, we propose a hierarchical prompt strategy named Hierarchical Prompt Tuning (\textbf{HiPTune}). HiPTune is designed to explore multiple classification criteria across different semantic spaces and adaptively integrate the most appropriate criterion for each sample. Specifically, following the hierarchical structure of our dataset (as shown the Fig.~\ref{fig_data} (d)), we first construct a Visual Prompt Tree (\textbf{VP-Tree}) as a carrier of multiple classification criteria. 
The visual prompts are organized in a tree structure, progressing from coarse to fine, where each level represents a different granularity of classification criteria, and each leaf node corresponds to a specific criterion that is robust enough to distinguish certain fake samples from real faces. 
Shown as Fig.~\ref{fig_Motivation} (b), our HiPTune aims to utilize coarse-to-fine criteria to achieve a more adequate classification. This design endows the prompts with both global discriminative power and local specialization: top-level prompts enhance recall for unknown forgeries, while lower-level prompts improve fine-grained recognition of known categories. Moreover, to enrich the semantic information of these prompts, we introduce a Dynamic Prompt Integration (\textbf{DPI}) module, which employs a self-attention structure to adaptively integrate prompts from different levels and map them into the text embedding dimension. This cross-modal integration strategy broadens the semantic coverage of classification criteria, ensuring classification accuracy from multiple dimensions. Finally, we ultilize a Adaptive Prompt Pruning (APP) mechanism that calculate the cross-attention scores between the input samples and the prompts, guiding the model to adaptively select the appropriate classification criteria at different levels for each sample. 
To summarize, our contributions are as follows:
\begin{itemize}
	\item
	We built on the identity-consistent design of UniAttackData~\cite{fang2024unified} to create \textbf{UniAttackData+}, the largest UAD dataset to date, featuring $54$ attack types and over $679,000$ videos from $2,875$ participants. Collected from $2,875$ participants across three ethnic groups, the dataset includes 18,250 real videos under diverse conditions, with at least 30 frames sampled per video for training and evaluation.
	\item
	 We propose \textbf{HiPTune}, a hierarchical prompt tuning strategy that leverages a \textbf{VP-Tree} and a \textbf{APP} to organize multiple classification criteria from coarse to fine, enabling adaptive selection for each sample. By integrating a \textbf{DPI} module with self-attention, HiPTune enhances semantic richness and cross-modal alignment, allowing the model to balance global generalization and local discrimination for improved face attack detection.
	\item
	We propose four intra-dataset protocols and nine cross-dataset benchmarks to promote the development of UAD community.
\end{itemize}

\begin{table}[]
	\centering
	\caption{Comparison of existing PAD and FFD datasets. The scale of our UniAttackData+ is far more than the types in the existing PAD and FFD datasets. `(I)' indicates that the dataset only contains images.}
	\scalebox{1.2}{
		\begin{tabular}{|l|c|c|}
			\hline
			Dataset                 & \# Attack Types     & \# Videos        \\ \hline \hline
			CASIA-FASD~\cite{Zhang2012A}                            & phy.2                    & 600              \\ \hline
			Replay-Attack~\cite{Chingovska_BIOSIG-2012}      & phy.3                   & 1,200            \\ \hline
			MSU-MFSD~\cite{wen2015face}                & phy.3                    & 440                           \\ \hline
			HKBU-MARs V2~\cite{liu20163d}                & phy.2 & 1,008  \\ \hline 
			OULU-NPU~\cite{Boulkenafet2017OULU}           & phy.4                    & 5,940             \\ \hline
			SiW~\cite{Liu2018Learning}                         & phy.6 & 4,620   \\ \hline
			Rose-Youtu~\cite{li2018unsupervised}       & phy.4 & 4,000  \\ \hline
			CelebA-Spoof~\cite{CelebA-Spoof}            & phy.4 & 625,537(I)  \\ \hline
			3DMask~\cite{yu2020fas}                      & phy.3 & 1,152   \\ \hline
			SiW-Mv2~\cite{guo2022multi}                  & phy.14                   & 1,700            \\ \hline
			HiFiMask~\cite{liu2022contrastive}                & phy.3                    & 54,600           \\ \hline  
			WMCA~\cite{george2019biometric}              & phy.7 & 6,716  \\ \hline
			CASIA-SURF~\cite{zhang2020casia}              & phy.6                    & 21,000           \\ \hline
			PADISI-Face~\cite{rostami2021detection}  & phy.9 & 121,740(I)  \\ \hline    \hline
			DFD~\cite{dufour2019deepfakes}     & digi.5 & 3,431  \\ \hline
			DFDC~\cite{dolhansky2019deepfake}               & digi.8                   & 128,064          \\ \hline
			FaceForensics++~\cite{rossler2019faceforensics++}         & digi.4                   & 5,000            \\ \hline
			Celeb-DF~\cite{li2020celeb}                & digi.1 & 6,229  \\ \hline
			DeeperForensics-1.0~\cite{Jiang_2020_CVPR}     & digi.1                   & 60,000           \\ \hline
			FFIW~\cite{Zhou_2021_CVPR}       & digi.3   & 20,000  \\ \hline
			ForgeryNet~\cite{he2021forgerynet}              & digi.15                  & 221,247          \\ \hline \hline
			\rowcolor{gray!30}
			\textbf{UniAttackData+} & \textbf{phy.14, digi.40} & \textbf{697,347} \\ \hline
		\end{tabular}
	}
	\label{tab_DataCompare}
\end{table}

\section{Related Works}
\subsection{Face Attack Datasets}
\subsubsection{PAD Datasets}
CASIA-FASD~\cite{Zhang2012A} comprises 50 subjects, with each subject having 12 videos (3 genuine and 9 fake). The videos are captured in three different imaging qualities: low, normal, and high. Two types of face attacks are included in the dataset: photo and video attacks. Replay-Attack~\cite{Chingovska_BIOSIG-2012}, comprises short video recordings of both real-access and attack attempts on 50 clients in 1,200 videos. It includes three types of attacks: two kinds of replayed faces and one kind of printed face. MSU-MFSD~\cite{wen2015face} comprises 440 videos across 35 subjects, employing two mobile devices to investigate three types of spoof attacks: 1 printed photo, and 2 video replay. HKBU-MARs V2~\cite{liu20163d} contains 1,008 videos from 12 subjects and 2 types of 3D masks: Thatsmyface and REAL-F masks. OULU-NPU~\cite{Boulkenafet2017OULU} contains 5,940 high-resolution videos from 55 subjects, including 1,980 real-access and 3,960 attack videos. Captured with six mobile front cameras over three sessions, it features varying lighting and backgrounds, and includes 2 printers and 2 displays. SiW~\cite{Liu2018Learning} contains 4,620 30-fps videos from 165 subjects, where each subject has 6 spoof types: 2 print, and 4 replay. Live videos are recorded across four sessions featuring variations in head movement, facial expression, and illumination. Rose-Youtu~\cite{li2018unsupervised} contains over 4,000 videos of 25 subjects captured under varied illumination, camera models, and three spoofing attacks (printed paper, video replay, and masking). CelebA-Spoof~\cite{CelebA-Spoof} is a richly annotated dataset with 625,537 images from 10,177 subjects. It includes four types of spoofing attacks, \ie, print, paper-cut, replay, and 3D mask, captured under four illumination conditions and two environmental settings, \ie, indoor and outdoor. SiW-Mv2~\cite{guo2022multi} has 785 videos from 493 live subjects, and 915 spoof videos from 600 subjects. Among these spoof videos, it has 14 spoof attack types, spanning from typical 2D spoof attacks (\eg, print and replay), various masks, different makeups, and physical material coverings. 3DMask~\cite{yu2020fas} contains 288 live videos and 864 3D mask videos from 48 subjects. Uniquely, it includes outdoor scenes with challenging lighting and considers three mask decoration types. HiFiMask~\cite{liu2022contrastive} is a realistic masks dataset, with a total of 54,600 videos recorded from 75 subjects with 3 kinds of high-fidelity masks. 

WMCA~\cite{george2019biometric} includes a diverse range of 2D and 3D presentation attacks that can be grouped into 7 categories. It provides 6,716 videos with multi-channel data, \ie, color, depth, near-infrared, and thermal. CASIA-SURF~\cite{zhang2020casia} is the first large-scale multimodal PAD dataset, which consists of 1,000 subjects with 21,000 videos, and each sample has 3 modalities (\ie, RGB, Depth, and IR) and 6 print attacks. PADISI-Face~\cite{rostami2021detection} is a realistic and challenging dataset, and includes 121,740 frames from 9 attack types. 

\subsubsection{FFD Datasets}
DFD~\cite{dufour2019deepfakes} dataset, developed by Google in partnership with Jigsaw, contains 3,431 high-quality videos of actors speaking in 5 deepfake approaches. DFDC~\cite{dolhansky2019deepfake} is the largest publicly available face swap video dataset, with 128,064 video clips, produced with 8 Deepfake, GAN-based, and non-learned methods. FaceForensics++~\cite{rossler2019faceforensics++} contains 4,000 videos, and manipulations created with 4 methods, namely, Face2Face, FaceSwap, DeepFakes, and NeuralTextures. Celeb-DF~\cite{li2020celeb} contains 590 real videos, and 5,639 high-quality DeepFake videos of celebrities generated with a synthesis algorithm. DeeperForensics-1.0~\cite{Jiang_2020_CVPR} contains 60,000 videos constituted by a total of 17.6 million frames, where the 10,000 fake videos are generated by a proposed face swapping framework. FFIW~\cite{Zhou_2021_CVPR} comprises 10,000 real and high-quality forgery videos, with an average of three human faces in each frame, and only a small subset has been manipulated with one of three face swapping methods. ForgeryNet~\cite{he2021forgerynet} is the largest publicly available deepfake dataset, containing 2.9 million images and 221,247 videos with unified annotations across image- and video-level tasks. It supports four major tasks, \ie, image and video forgery classification, spatial forgery localization, and temporal forgery localization, spanning 15 manipulation methods and 36 perturbations. 

\subsubsection{UAD Datasets}
GrandFake~\cite{deb2023unified} is the first UAD dataset developed to resolve the dilemma wherein a single model struggles to handle all three categories of attacks concurrently. It consists of 25 face attacks from 3 attack categories, \ie, Adversarial Faces, Digital Manipulation, Physical Spoofs. JFSFDB~\cite{yu2024benchmarking} dataset consists of 9 existing datasets, including 6 PAD datasets and 3 FFD datasets. However, the limited diversity of attack types and data volume in the GrandFake and JFSFDB datasets, coupled with the absence of identities shared across all attack types, constrains the model's generalization and makes it prone to overfitting identity-specific features. Based on the CASIA-SURF CeFA dataset~\cite{liu2021casia}, UniAttackData~\cite{fang2024unified} introduces 12 types of digital forgery on real faces, including 6 adversarial attacks and 6 attribute editing attacks. Although UniAttackData ensures that each identity includes all types of attacks, it is limited in the diversity of physical and digital forgeries, lacking high-fidelity mask attacks and advanced generative techniques.

\subsection{Face Attack Detection Methods}
UniFAD~\cite{deb2023unified} is the first UAD method that combines k-means clustering and multi-task learning to jointly learn representations for coherent attack types while handling uncorrelated ones separately. UniAttackDetection~\cite{fang2024unified} is a vision-language-based UAD method that integrates teacher-student prompts, unified knowledge mining, and sample-level prompt interaction to effectively capture both general and sample-specific attack semantics. La-SoftMoE~\cite{zou2024softmoe} leverages a flexible Mixture of Experts with self-adaptive weighting to effectively handle sparse and diverse attack data across complex feature spaces. MoAE-CR~\cite{chen2025mixture} combines Mixture-of-Attack-Experts for fine-grained feature extraction with class-aware regularization modules to enhance intra-class compactness and inter-class separability for improved detection of diverse attacks. FA$^{3}$-CLIP~\cite{li2025fa} combines spatial and frequency features through dual-stream fusion and attack-agnostic prompt learning to generate generalized live/fake representations, enabling robust detection across diverse physical and digital attacks.

\begin{table*}[htbp]
	\centering
	\caption{Comparison of existing UAD datasets. UniAttackData+ surpasses all other datasets in scale and diversity. `V' and `I' indicate video- and image-based counts, respectively. The numbers following each attack type in the `\# Catagories' row represent the number of algorithms included in that category.}
	\begin{tabular}{|c|c|c|c|cc|cc|}
		\hline
		\multirow{3}{*}{\textit{Dataset}}                 & \multirow{3}{*}{\begin{tabular}[c]{@{}c@{}}Attack Type\\ (each ID)\end{tabular}} & \multirow{3}{*}{\# Datasets / Data}                                                                                      & \multirow{3}{*}{\# ID}          & \multicolumn{2}{c|}{\multirow{2}{*}{Physical Attacks}}                                                                         & \multicolumn{2}{c|}{\multirow{2}{*}{Digital Attacks}}                            \\
		&                                                                                  &                                                                                                                          &                                 & \multicolumn{2}{c|}{}                                                                                                          & \multicolumn{2}{c|}{}                                                            \\ \cline{5-8} 
		&                                                                                  &                                                                                                                          &                                 & \multicolumn{1}{c|}{Dataset Name}                                                               & No.                          & \multicolumn{1}{c|}{\# Categories}                 & No.                         \\ \hline \hline
		\multirow{2}{*}{\textit{GrandFake}}               & \multirow{2}{*}{Incomplete}                                                      & \multirow{2}{*}{\begin{tabular}[c]{@{}c@{}}6 sets: 789,412 (I)\\ (Live: 341738, Fake: 447674)\end{tabular}}              & \multirow{2}{*}{96,817}         & \multicolumn{1}{c|}{\multirow{2}{*}{SiW-M}}                                                     & \multirow{2}{*}{128,112 (I)} & \multicolumn{1}{c|}{Adv (6)}                       & 116,641 (I)                 \\ \cline{7-8} 
		&                                                                                  &                                                                                                                          &                                 & \multicolumn{1}{c|}{}                                                                           &                              & \multicolumn{1}{c|}{DeepFake (6)}                  & 202,921 (I)                 \\ \hline
		\multirow{6}{*}{\textit{JFSFDB}}                  & \multirow{6}{*}{Incomplete}                                                      & \multirow{6}{*}{\begin{tabular}[c]{@{}c@{}}9 sets: 27,172 (V)\\ (Live: 5,650, Fake: 21,522)\end{tabular}}                & \multirow{6}{*}{356}            & \multicolumn{1}{c|}{SiW}                                                                        & 3,173 (V)                    & \multicolumn{1}{c|}{\multirow{6}{*}{DeepFake (4)}} & \multirow{6}{*}{13,752 (V)} \\ \cline{5-6}
		&                                                                                  &                                                                                                                          &                                 & \multicolumn{1}{c|}{3DMAD}                                                                      & 85 (V)                       & \multicolumn{1}{c|}{}                              &                             \\ \cline{5-6}
		&                                                                                  &                                                                                                                          &                                 & \multicolumn{1}{c|}{HKBU}                                                                       & 588 (V)                      & \multicolumn{1}{c|}{}                              &                             \\ \cline{5-6}
		&                                                                                  &                                                                                                                          &                                 & \multicolumn{1}{c|}{MSU}                                                                        & 210 (V)                      & \multicolumn{1}{c|}{}                              &                             \\ \cline{5-6}
		&                                                                                  &                                                                                                                          &                                 & \multicolumn{1}{c|}{3DMask}                                                                     & 864 (V)                      & \multicolumn{1}{c|}{}                              &                             \\ \cline{5-6}
		&                                                                                  &                                                                                                                          &                                 & \multicolumn{1}{c|}{ROSE}                                                                       & 2,850 (V)                    & \multicolumn{1}{c|}{}                              &                             \\ \hline
		\multirow{2}{*}{\textit{UniAttackData}}           & \multirow{2}{*}{Complete}                                                        & \multirow{2}{*}{\begin{tabular}[c]{@{}c@{}}1 set: 28,706 (V)\\ (Live: 1,800, Fake: 26,906)\end{tabular}}                 & \multirow{2}{*}{1,800}          & \multicolumn{1}{c|}{\multirow{2}{*}{\begin{tabular}[c]{@{}c@{}}CASIA-SURF\\ CeFA\end{tabular}}} & \multirow{2}{*}{5,400 (V)}   & \multicolumn{1}{c|}{Adv (6)}                       & 10,706 (V)                  \\ \cline{7-8} 
		&                                                                                  &                                                                                                                          &                                 & \multicolumn{1}{c|}{}                                                                           &                              & \multicolumn{1}{c|}{DeepFake (6)}                  & 10,800 (V)                  \\ \hline
		\multirow{3}{*}{\textit{\textbf{UniAttackData+}}} & \multirow{3}{*}{\textbf{Complete}}                                               & \multirow{3}{*}{\textbf{\begin{tabular}[c]{@{}c@{}}3 set: 697,347 (V)\\ (Live: 18,250, Fake: 679,097)\end{tabular}}} & \multirow{3}{*}{\textbf{2,875}} & \multicolumn{1}{c|}{\textbf{CASIA-SURF}}                                                        & \textbf{6,000 (V)}           & \multicolumn{1}{c|}{\textbf{Adv (16)}}             & \textbf{266,576 (V)}        \\ \cline{5-8} 
		&                                                                                  &                                                                                                                          &                                 & \multicolumn{1}{c|}{\textbf{CeFA}}                                                              & \textbf{9,000 (V)}           & \multicolumn{1}{c|}{\textbf{DeepFake (17)}}        & \textbf{242,859 (V)}        \\ \cline{5-8} 
		&                                                                                  &                                                                                                                          &                                 &
	 \multicolumn{1}{c|}{\textbf{HiFiMask}}                                                          & \textbf{40,950 (V)}          & \multicolumn{1}{c|}{\textbf{Generation (7)}}      & \textbf{113,712 (V)}        \\ \hline
	\end{tabular}
	\label{tab_DataCount}
\end{table*}

\section{UniAttackDataPlus}
\subsection{Advantages of UniAttackData+}
Our UniAttackData+ addresses three critical issues in current UAD datasets, \ie, insufficient sample size, limited attack coverage, and lack of interpretability, providing a comprehensive and valuable benchmark for the UAD community.

\textbf{Scale.} UniAttackData+ includes $2,875$ identities spanning three ethnicities, totaling $697,347$ video segments, with at least $25$ frames per video. Of these, $18,250$ are authentic face videos, 55,950 are physical attack videos, and $623,147$ are digital attack videos. Detailed data is presented in Tab.~\ref{tab_DataCount} and Tab.~\ref{tab_DataCompare}. Our UniAttackData+ shows its superiority in separate PAD and FFD datasets, and also in the existing UAD datasets. It ensures a comprehensive and complete learning space, guaranteeing the effectiveness of detection models from a data perspective.

\textbf{Attack Types.} UniAttackData+ comprises $54$ attack types in total, including $14$ physical attacks and $40$ digital attacks, covering all known attack types in the field. This diversity ensures the model does not overfit to a specific attack type, thereby enhancing its generalization ability. All the attack methods included in our dataset were proposed after 2021, incorporating the latest digital generation attacks. UniAttackData+ promotes real-world generalizability by significantly expanding attack diversity, enabling more comprehensive training and reducing model susceptibility to novel advanced attacks.

\textbf{Interpretability.} Existing UAD datasets merely merge different datasets without any correlation between the data, attack methods, and identities. This simplistic combination hinders the model from focusing on genuine forgery cues and greatly reduces model robustness in real-world applications. Unlike previous datasets, UniAttackData+ maintains the identity consistency of UniAttackData~\cite{fang2024unified} and further organizes all attack types into a hierarchical taxonomy, enabling each sample to be labeled from coarse to fine levels. The hierarchical labeling scheme offers a richer experimental and analytical framework, facilitating more interpretable and fine-grained evaluation of detection models.

\subsection{Collection of UniAttackData+}
\textbf{Live Face source.} The live faces in the UniAttackData+, enhancing diversity in terms of face identity, angle, expression, and ethnicity, are sourced from three public datasets: CASIA-SURF~\cite{zhang2020casia}, which includes $1,000$ subjects; CASIA-SURF CeFA~\cite{liu2021casia} (CeFA), comprising 1,800 subjects from three ethnicities, \ie, African, East Asian, and Central Asian, with $600$ subjects each ethnicity; and CASIA-SURF HiFiMask~\cite{liu2022contrastive} (HiFiMask), which contains $75$ subjects. These varied sources significantly enhance the dataset’s diversity across multiple dimensions. We merge three existing datasets for the live face data to obtain $2,875$ distinct identities. CASIA-SURF and CeFA use the RealSense SR300 to capture head movements and distance variations. HiFiMask simulates real-world lighting with white, green, and tri-color cycles, plus six-directional supplementary lights.

\textbf{Preprocessing.} For dataset preprocessing, we apply MTCNN~\cite{timesler2020facenet_pytorch} for face detection and cropping on the real face data from the original datasets, with the scale parameter set to 1.3. We then manually assign class labels, attack type labels, and hierarchical labels to facilitate research.

\subsection{Spoofing Approach}
\begin{figure*}[t]
	\centering
	\includegraphics[width=0.99\linewidth]{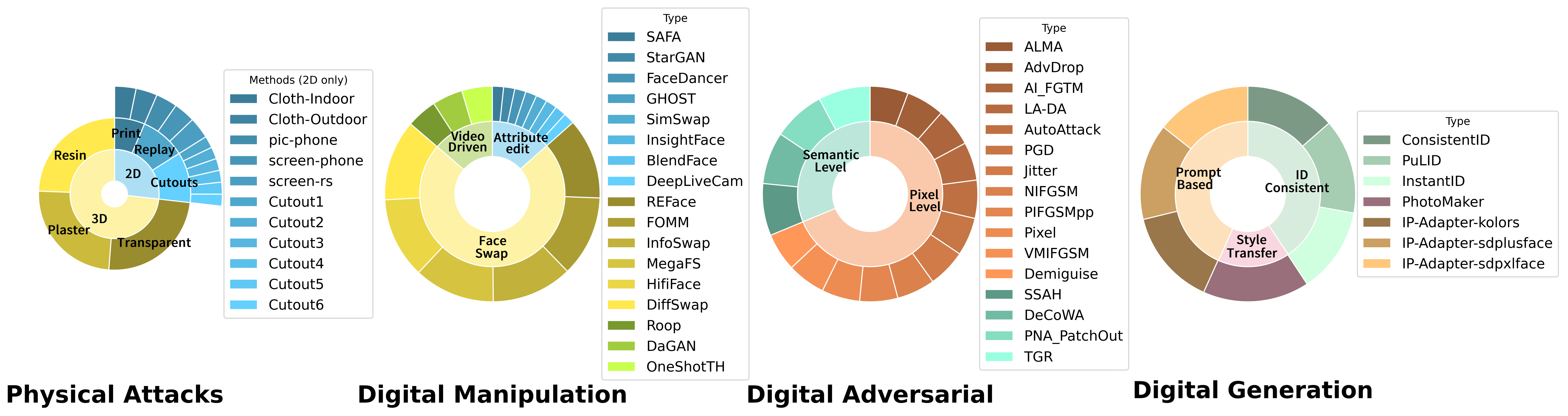}
	\caption{This diagram categorizes attacks into four distinct types: Physical Attacks, Digital Manipulation, Digital Adversarial and Digital Generation. Physical Attacks comprise 14 subtypes, Digital Manipulation includes 17 subtypes, the Digital Adversarial group contains 16 subtypes, and the Digital Generation group has 7 subtypes. We can clearly see two advantages of datasets containing forgery methods: (2) rich and diverse types of attacks; (2) Determine the proportion based on the commonness and progressiveness of attack types.}
	\label{fig:proportion_statistics}
\end{figure*}

To ensure diversity in physical attack methods within the UniAttackData+ dataset, we incorporate $14$ physical face spoofing techniques, including $11$ 2D presentation attacks and $3$ 3D mask attacks. See from the Tab.~\ref{fig:proportion_statistics}, The 2D attacks consist of $6$ paper-cutout variations and $2$ print attack types, as well as $3$ video-replay attacks. For 3D mask attacks, UniAttackData+ includes high-fidelity masks made from $3$ different materials: transparent, plaster, and resin. These masks are recorded under varied lighting conditions and with multiple imaging devices for added variation.

\textbf{2D Attacks.} We include two kinds of print attacks recorded in the outdoor and indoor environments. Moreover, three kinds of screen replay attacks from different devices are utilized. For the cutouts, the printed flat or curved face images will be cut into eyes, nose, mouth areas, or their combinations, generating 6 different attack ways. Specifically,  the six ways are: (1) One person holds his/her flat face photo where eye regions are cut; (2) One person holds his/her curved face photo where eye regions are cut; (3) One person holds his/her flat face photo where eye and nose regions are cut; (4) One person holds his/her curved face photo where eye and nose regions are cut; (5) One person holds his/her flat face photo where eye, nose, and mouth regions are cut; (6) One person his/her curved face photo where eye, nose, and mouth regions are cut.

\textbf{3D Attacks.} For the 3D attacks, we construct three kinds of masks with different materials: transparent, plaster, and resin. All masks are worn on a real person's face and uniformly dressed to avoid the algorithm looking for clues outside the head area.

\subsection{Forgery Approach}
\subsubsection{Digital Manipulation Attacks}
As shown in Tab.~\ref{fig:proportion_statistics}, Digital Manipulation Attacks in UniAttackData+ encompasses $17$ specific methods grouped into three types of attacks: (1) Face Swapping~\cite{rosberg2023facedancer,9851423ghost,chen2020simswap, ren2023pbidrinsightface,shiohara2023blendface,baliah2024realistic,siarohin2021motion, gao2021information,zhu2021one, wang2021hififace,zhao2023diffswap}, where identity features of a source face, $I_{\text{source}}$, are transferred into a target face, $I_{\text{target}}$, producing a composite image, $I_{\text{composite}} = f(I_{\text{source}}, I_{\text{target}})$. (2) Facial Attribute Editing~\cite{wang2021safa, choi2018stargan}, which alters specific attributes of a face, such as age, expression, or gender, expressed as $I' = g(I, A_{\text{target}})$, where $A_{\text{target}}$ denotes the desired attribute. (3) Video-Driven Facial Animation~\cite{hong2022depth, wang2021one}, which generates facial movements in static images using motion parameters \(M\) from a driving source, represented by \(I_{\text{animated}} = h(I, M)\).

\textbf{Face-Swap.} FaceDancer~\cite{rosberg2023facedancer} utilizes Adaptive Feature Fusion Attention (AFFA) and Interpreted Feature Similarity Regularization (IFSR) to achieve seamless face-swapping with high-fidelity identity transfer and pose preservation. GHOST~\cite{9851423ghost} introduces architectural enhancements like an eye-based loss function, face mask smoothing, and stabilization techniques to achieve high-fidelity image-to-image and image-to-video face swapping with improved identity preservation and gaze alignment. SimSwap~\cite{chen2020simswap} employs an ID Injection Module (IIM) for feature-level identity transfer and a Weak Feature Matching Loss to preserve target facial attributes, enabling high-fidelity and generalized face swapping. BlendFace~\cite{shiohara2023blendface} introduces a novel identity encoder trained on blended images to mitigate attribute biases, enabling better identity-attribute disentanglement for improved face-swapping while maintaining competitive performance. Deep-Live-Cam~\cite{deeplivecam} provides tools for animating custom characters and modeling tasks, with built-in safeguards to prevent processing of inappropriate content, emphasizing ethical and lawful use, user responsibility, and adherence to legal requirements, including potential watermarking or project suspension if necessary. REFace~\cite{baliah2024realistic} leverages diffusion models to frame face-swapping as a self-supervised inpainting task, incorporating DDIM sampling, CLIP feature disentanglement, and mask shuffling for enhanced identity transfer, realism, and versatility, including head and accessory swapping. FOMM~\cite{siarohin2021motion} introduces unsupervised motion representations for animating articulated objects by identifying and tracking distinct parts, using region-based descriptions to disentangle shape and pose, and modeling global motion with affine transformations for improved animation fidelity. InfoSwap~\cite{gao2021information} uses an information bottleneck and identity contrastive loss to enhance identity disentanglement, generating high-quality, identity-discriminative swapped faces. MegaFS~\cite{zhu2021one} introduces a Hierarchical Representation Face Encoder (HieRFE) and a Face Transfer Module (FTM) for high-quality megapixel-level face swapping, leveraging StyleGAN2 for stable synthesis and low memory usage. HifiFace~\cite{wang2021hififace} employs 3D shape-aware identity with geometric supervision and a Semantic Facial Fusion module for adaptive blending, achieving superior identity preservation and photo-realistic face-swapping results. DiffSwap~\cite{zhao2023diffswap} reframes face swapping as a conditional inpainting task using diffusion models, introducing a midpoint estimation method for efficient training and leveraging 3D-aware landmarks for controllable, high-fidelity, and shape-preserving face swapping. Roop~\cite{roop2023} takes a video and replaces the face in it with a face of your choice. You only need one image of the desired face.

\textbf{Attribute-edit.} SAFA~\cite{wang2021safa} employs 3D morphable models (3DMM) for facial structure modeling and inpainting techniques for occlusion recovery, enabling realistic and structure-aware face animation even for significantly deviated poses. StarGAN~\cite{choi2018stargan} employs a unified model architecture that enables multi-domain image-to-image translation within a single network, achieving superior scalability, robustness, and image quality compared to existing models.

\textbf{Video-Driven.} DaGAN~\cite{hong2022depth} leverages self-supervised depth learning to recover dense 3D facial geometry, using it for keypoint estimation and 3D-aware cross-modal attention to guide motion field generation, enabling realistic and accurate talking head video synthesis. OneShotTH~\cite{wang2021one} uses a novel keypoint representation to decompose identity and motion information unsupervisedly, enabling high-quality talking-head video synthesis with reduced bandwidth and interactive head rotation for enhanced video conferencing experiences.

\begin{table*}[]
	\centering
	\caption{Sample statistics for each protocol on the UniAttackData+ dataset, including train, dev, and test sets, where each column represents the number of samples for live faces or fake types. `\#' indicates the number of videos.}
	\resizebox*{1.0 \linewidth}{!}{
		\begin{tabular}{|c|c|c|cc|ccc|cc|ccc|c|}
			\hline
			\multirow{2}{*}{\textbf{Protocol}} & \multirow{2}{*}{\textbf{Subset}} & \multirow{2}{*}{\textbf{\#Live}} & \multicolumn{2}{c|}{\textbf{Physical Attacks}}     & \multicolumn{3}{c|}{\textbf{Digtal Manipulation}}                                                                                                                                                                                                     & \multicolumn{2}{c|}{\textbf{Digital Adversarial}}                                                                                                             & \multicolumn{3}{c|}{\textbf{Digital Generation}}                                                                                                                                                                                                          & \multirow{2}{*}{\textbf{\#All}} \\ \cline{4-13}
			&                                  &                                  & \multicolumn{1}{c|}{\textbf{\#2D}} & \textbf{\#3D} & \multicolumn{1}{c|}{\textbf{\begin{tabular}[c]{@{}c@{}}\#Attribute-\\ Edit\end{tabular}}} & \multicolumn{1}{c|}{\textbf{\begin{tabular}[c]{@{}c@{}}\#Face-\\ Swap\end{tabular}}} & \textbf{\begin{tabular}[c]{@{}c@{}}\#Video-\\ Driven\end{tabular}} & \multicolumn{1}{c|}{\textbf{\begin{tabular}[c]{@{}c@{}}\#Pixel-\\ Level\end{tabular}}} & \textbf{\begin{tabular}[c]{@{}c@{}}\#Semantic-\\ Level\end{tabular}} & \multicolumn{1}{c|}{\textbf{\begin{tabular}[c]{@{}c@{}}\#ID-\\ Consistent\end{tabular}}} & \multicolumn{1}{c|}{\textbf{\begin{tabular}[c]{@{}c@{}}\#Style-\\ Transfer\end{tabular}}} & \textbf{\begin{tabular}[c]{@{}c@{}}\#Prompt-\\ Based\end{tabular}} &                                 \\ \hline
			\multirow{3}{*}{P1}                & train                            & 9,125                            & \multicolumn{1}{c|}{7,500}         & 20,475        & \multicolumn{1}{c|}{12,330}                                                               & \multicolumn{1}{c|}{88,439}                                                          & 16,661                                                             & \multicolumn{1}{c|}{91,636}                                                            & 41,653                                                               & \multicolumn{1}{c|}{23,161}                                                              & \multicolumn{1}{c|}{9,096}                                                                & 24,600                                                             & 355,676                         \\ \cline{2-14} 
			& dev                              & 3,650                            & \multicolumn{1}{c|}{3,000}         & 8,190         & \multicolumn{1}{c|}{6,532}                                                                & \multicolumn{1}{c|}{35,376}                                                          & 6,664                                                              & \multicolumn{1}{c|}{36,654}                                                            & 16,661                                                               & \multicolumn{1}{c|}{9,264}                                                               & \multicolumn{1}{c|}{3,638}                                                                & 9,840                                                              & 139,469                         \\ \cline{2-14} 
			& text                             & 5,475                            & \multicolumn{1}{c|}{4,500}         & 12,285        & \multicolumn{1}{c|}{13,797}                                                               & \multicolumn{1}{c|}{53,063}                                                          & 9,997                                                              & \multicolumn{1}{c|}{54,981}                                                            & 24,991                                                               & \multicolumn{1}{c|}{13,896}                                                              & \multicolumn{1}{c|}{5,458}                                                                & 14,759                                                             & 213,202                         \\ \hline
			\multirow{3}{*}{P2}                & train                            & 13,650                           & \multicolumn{1}{c|}{5,880}         & 19,110        & \multicolumn{1}{c|}{11,663}                                                               & \multicolumn{1}{c|}{89,522}                                                          & 11,663                                                             & \multicolumn{1}{c|}{81,639}                                                            & 34,988                                                               & \multicolumn{1}{c|}{21,078}                                                              & \multicolumn{1}{c|}{9,096}                                                                & 23,026                                                             & 321,315                         \\ \cline{2-14} 
			& dev                              & 1,000                            & \multicolumn{1}{c|}{2,520}         & 8,190         & \multicolumn{1}{c|}{4,998}                                                                & \multicolumn{1}{c|}{38,366}                                                          & 4,998                                                              & \multicolumn{1}{c|}{34,988}                                                            & 14,995                                                               & \multicolumn{1}{c|}{9,033}                                                               & \multicolumn{1}{c|}{3,638}                                                                & 9,868                                                              & 132,594                         \\ \cline{2-14} 
			& text                             & 3,600                            & \multicolumn{1}{c|}{6,600}         & 13,650        & \multicolumn{1}{c|}{15,998}                                                               & \multicolumn{1}{c|}{48,990}                                                          & 16,661                                                             & \multicolumn{1}{c|}{49,983}                                                            & 33,322                                                               & \multicolumn{1}{c|}{16,210}                                                              & \multicolumn{1}{c|}{5,458}                                                                & 16,301                                                             & 226,773                         \\ \hline
			\multirow{3}{*}{P3}                & train                            & 13,650                           & \multicolumn{1}{c|}{10,500}        & -             & \multicolumn{1}{c|}{22,861}                                                               & \multicolumn{1}{c|}{123,815}                                                         & 23,325                                                             & \multicolumn{1}{c|}{128,290}                                                           & 58,314                                                               & \multicolumn{1}{c|}{-}                                                                   & \multicolumn{1}{c|}{-}                                                                    & -                                                                  & 380,755                         \\ \cline{2-14} 
			& dev                              & 1,000                            & \multicolumn{1}{c|}{4,500}         & -             & \multicolumn{1}{c|}{9,798}                                                                & \multicolumn{1}{c|}{53,063}                                                          & 9,997                                                              & \multicolumn{1}{c|}{54,981}                                                            & 24,991                                                               & \multicolumn{1}{c|}{-}                                                                   & \multicolumn{1}{c|}{-}                                                                    & -                                                                  & 158,330                         \\ \cline{2-14} 
			& text                             & 3,600                            & \multicolumn{1}{c|}{-}             & 40,950        & \multicolumn{1}{c|}{-}                                                                    & \multicolumn{1}{c|}{-}                                                               & -                                                                  & \multicolumn{1}{c|}{-}                                                                 & -                                                                    & \multicolumn{1}{c|}{46,321}                                                              & \multicolumn{1}{c|}{18,192}                                                               & 49,199                                                             & 158,262                         \\ \hline
		\end{tabular}
	}
	\label{tab:StatisticsProtocols}
\end{table*}

\subsubsection{Digital Adversarial Attacks}
UniAttackData+ includes $16$ attacks designed to mislead machine learning models through subtle, often imperceptible perturbations. These attacks are categorized into pixel-level attacks~\cite{rony2021augmented, duan2021advdrop, zou2022making, yan2022ila, croce2020reliable, madry2017towards, lin2024boosting, schwinn2023exploring, lin2019nesterov, gao2020patch, pomponi2022pixle, wang2021enhancing}, which modify raw image pixels to create adversarial examples, and semantic-level attacks~\cite{wang2021demiguise, luo2022frequency, wei2022towards, zhang2023transferable}, which alter high-level features to change model interpretations. Pixel-level attacks rely on precise, gradient-based adjustments, while semantic-level attacks manipulate intermediate representations or latent spaces, increasing their transferability and effectiveness.

\textbf{Pixel-Level.} ALMA~\cite{rony2021augmented} utilizes Augmented Lagrangian-based attack for efficient, minimally perturbed adversarial examples, combining generality and speed. AdvDrop~\cite{duan2021advdrop} introduces a novel adversarial attack by dropping imperceptible details from images instead of adding disturbances, challenging DNN robustness and evading existing defense mechanisms. AI-FGTM~\cite{zou2022making} improves adversarial example generation with dynamic step size, smaller kernels, and a tanh-based gradient method, achieving higher transferability, indistinguishability, and attack success rates than state-of-the-art methods. ILA-DA~\cite{yan2022ila} enhances adversarial example transferability by combining automated image transformations, reverse adversarial updates, and attack interpolation, significantly outperforming prior methods in both defended and undefended scenarios. AutoAttack~\cite{croce2020reliable} introduces an ensemble of parameter-free adversarial attacks, improving PGD with optimized step size and objective function to rigorously evaluate adversarial robustness, uncovering weaknesses in many published defenses. PGD~\cite{madry2017towards} explores adversarial robustness through robust optimization, introducing methods for training and attacking neural networks with universal security guarantees against first-order adversaries, significantly enhancing resistance to adversarial attacks. Jitter~\cite{schwinn2023exploring} proposes a novel loss function that improves attack efficiency and success rates by better exploring the perturbation space and addressing model confidence issues. NI-FGSM~\cite{lin2019nesterov} leverages Nesterov accelerated gradients and SIM, optimizing perturbations across scaled copies, to enhance the transferability of adversarial examples, achieving higher success rates on defense models. PIM++~\cite{gao2020patch} enhances adversarial transferability by using a patch-wise iterative method with an amplified step size, a projection kernel for gradient redistribution, and temperature adjustments to balance transferability and white-box attack performance. Pixel~\cite{pomponi2022pixle} introduces a black-box attack that rearranges a small number of pixels to achieve high attack success rates across models and datasets, requiring few iterations and minimal perceptual changes. VMI-FGSM~\cite{wang2021enhancing} enhances gradient-based adversarial attacks by stabilizing update directions with gradient variance, significantly improving transferability.

\textbf{Semantic-Level.} Demiguise Attack~\cite{wang2021demiguise} crafts unrestricted, photorealistic perturbations based on perceptual similarity, improving fooling rates, transferability, and robustness against defenses while maintaining human imperceptibility and simulating real-world conditions. SSAH~\cite{luo2022frequency} attacks semantic similarity in feature representations for cross-dataset adversarial generalization and employs a low-frequency constraint to generate imperceptible perturbations, achieving high transferability and perceptual similarity. DeCoWA~\cite{lin2024boosting} introduces Deformation-Constrained Warping (DeCoW) to enhance adversarial transferability across model genera by augmenting inputs with controlled elastic deformations, effectively attacking both CNNs and Transformers across various tasks. TGR~\cite{zhang2023transferable} reduces gradient variance in Vision Transformer (ViT) blocks token-wise, generating more transferable adversarial samples and improving attack performance. PNA-PatchOut~\cite{wei2022towards} combines Pay No Attention (PNA), skipping attention gradients, and PatchOut, optimizing random patch subsets, to enhance adversarial transferability across Vision Transformers (ViTs) and CNNs.

\subsubsection{Digital Generation Attacks}
UniAttackData+ comprises $7$ types leveraging diffusion models to produce highly realistic synthetic content. These attacks are divided into three categories: Identity-Consistent Generation, Style Transfer, and Prompt-Driven Generation. Identity-Consistent Generation~\cite{huang2024consistentid, guo2024pulid, wang2024instantid} seeks to maintain a subject's identity across various contexts, modeled as \( I_{\text{gen}} = G(I_{\text{id}}, z) \), where \( I_{\text{id}} \) is the identity input and \( z \) represents variations such as pose or expression. Style Transfer, as in PhotoMaker~\cite{li2024photomaker}, adapts identity features to different styles, represented as \( I_{\text{styled}} = G(I_{\text{id}}, \mathbf{S}_{\text{style}}) \), to create visually diverse outputs. Prompt-Driven Generation~\cite{ye2023ip} utilizes textual or image prompts to guide synthesis, expressed as \( I_{\text{prompted}} = G(P_{\text{text}}, P_{\text{img}}) \).

\textbf{ID-Consistent.} ConsistentID~\cite{huang2024consistentid} enhances personalized portrait generation with a multimodal facial prompt generator for fine-grained control and an ID-preservation network with facial attention localization, achieving superior ID consistency and precision from a single reference image. PuLID~\cite{guo2024pulid} introduces a tuning-free ID customization method for text-to-image generation, combining a Lightning T2I branch with contrastive alignment and ID losses to ensure high ID fidelity and minimal disruption to original image elements. InstantID~\cite{wang2024instantid} introduces a plug-and-play module with IdentityNet, leveraging strong semantic and weak spatial conditions to achieve high-fidelity, single-image-based identity personalization, seamlessly integrating with pre-trained diffusion models like SD1.5 and SDXL.

\textbf{Style-Transfer.} PhotoMaker~\cite{li2024photomaker} introduces a stacked ID embedding to encode multiple ID images into a unified representation, enabling efficient personalized text-to-image generation with high ID fidelity, text controllability, and speed, supported by an ID-oriented data construction pipeline.

\textbf{Prompt-Based.} IP-Adapter~\cite{ye2023ip} introduces a decoupled cross-attention mechanism to enable lightweight image prompting for text-to-image diffusion models, achieving multimodal generation with only 22M parameters, while maintaining compatibility with text prompts, structural controls, and custom models.

\subsection{Protocol and Statistic}
\textbf{Identity Division Protocol $P1$.} Simply merging existing PAD and FFD datasets can lead models to focus on identity-related noise rather than truly discriminative features. To mitigate this issue, we leverage the ID consistency provided by UniAttackData+. In this protocol, the model is trained on a selected subset of identities, while the remaining identities are reserved for validation and testing. Each identity includes all types of attacks to ensure comprehensive evaluation. 

\textbf{Cross-attack Methods Protocol $P2$.} In real-world deployment scenarios, a wide range of attack methods may emerge. Therefore, models must not only perform well on known attack types but also generalize effectively to unseen ones. This protocol is designed to evaluate a model’s generalization capability across diverse attack methods. All identities are used for training, while attack methods are split randomly: 50\% for training, 20\% for validation, and 30\% for testing. For example, in the case of digital attacks, the first seven types of digital face, \ie, swap attacks, are used for training, two types for validation, and four types for testing, enabling us to assess the model’s ability to handle unseen face-swap attacks.

\textbf{Cross-attack types Protocol $P3$.} In many cases, models are trained only on simpler attack types but are tested against more advanced and complex attack types. This protocol aims to assess the model's generalization ability from simpler to more sophisticated attacks. We trained the model using all identities, with 70\% of physical 2D attacks, digital editing, and adversarial attacks from each identity as the training set, and 30\% as the validation set. Physical 3D attacks and digital generative attacks were used as the test set to evaluate the model's generalization ability on unseen, more advanced attack types.

The sample statistics for each protocol are shown in Tab.~\ref{tab:StatisticsProtocols}. We have detailed the number of train, dev, and test sets for each protocol, and also listed the types of spoofing contained in the subsets and their quantities.

\section{Hierarchical Prompt Tuning (HiPTune)}
\begin{figure*}[!t]
	\centering
	\includegraphics[width=1\linewidth]{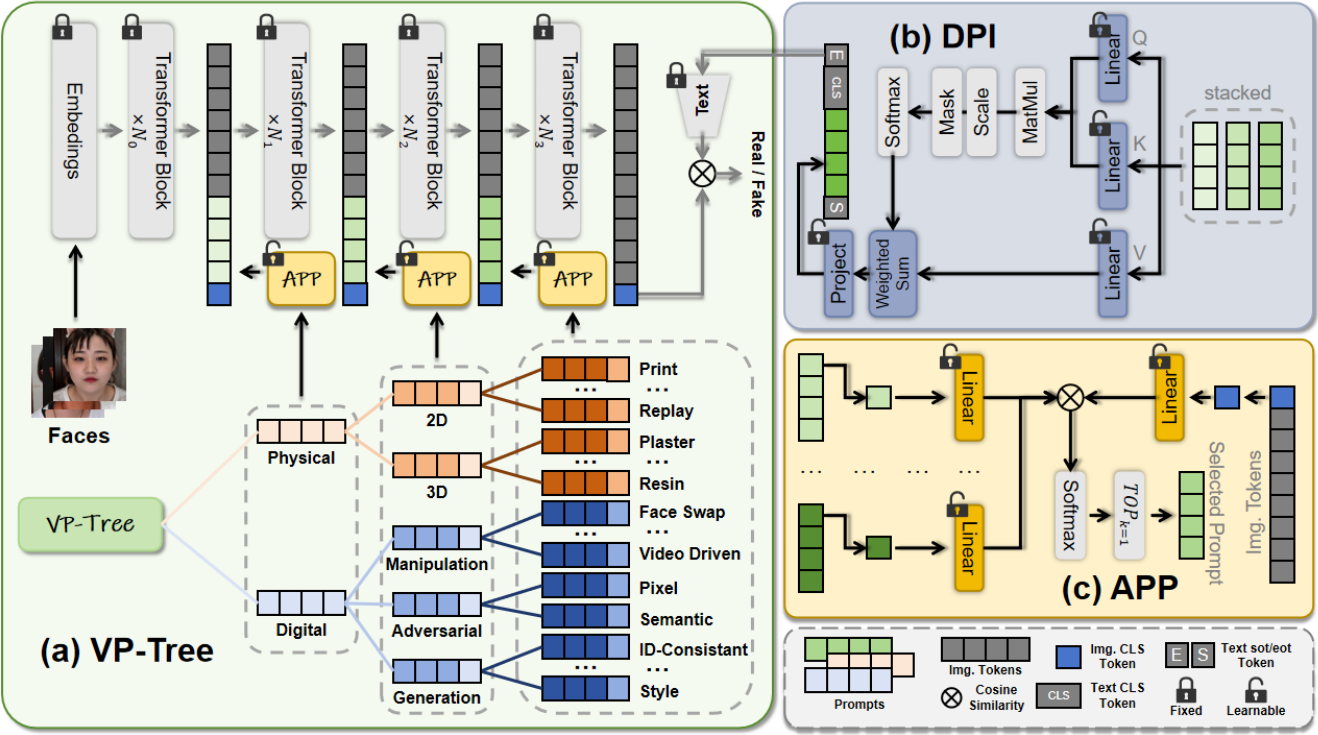}
	\caption{(a) The VP-Tree establishes a hierarchical prompt tree for all attack types, organized into three levels of grouping categories. It assigns a learnable prompt for each attack type and integrates the hierarchical prompts into different levels of the encoder. (b) The DPI aims to aggregate all candidate prompts from different levels of the VP-Tree and projects them into the text encoder for subsequent similarity computation. (c) The APP employs an adaptive prompt pruning mechanism to select the most promising prompt for each sample.}
	\label{fig_HiPTune}
\end{figure*}

\subsection{Preliminaries}
Vision-Language Models (VLMs) like CLIP~\cite{radford2021learning} utilize both text encoder $\mathcal{T}(\cdot)$ (like Transformer~\cite{vaswani2017attention}) and image encoder $\mathcal{I}(\cdot)$ (like ViT~\cite{dosovitskiy2020vit}), where the input is the image-text pair, providing a wide range of semantic information as supervision. The outputs of both encoders are then dot-multiplied to obtain the probability score $p$:
\begin{equation}
	p(y=i \mid \boldsymbol{x})=\frac{\exp \left(\cos \left(\boldsymbol{w}_{\boldsymbol{i}}, \boldsymbol{f}\right) / \tau\right)}{\sum_{j=1}^{K} \exp \left(\cos \left(\boldsymbol{w}_{\boldsymbol{j}}, \boldsymbol{f}\right) / \tau\right)},
\end{equation}
where $\left\{\boldsymbol{w}_{i}\right\}_{i=1}^{K}$ denotes the set of weight vectors generated by the text encoder, and $ \boldsymbol{f}$ represents the features extracted by the image encoder. $K$ is the number of classes, and $\tau$ refers to the temperature parameter learned by CLIP, $\cos(\cdot,\cdot)$ means the cosine similarity. Moreover, CoOp~\cite{zhou2022learning} explores prompt engineering in the CLIP, constructing learnable prompts $\boldsymbol{t}=[\mathrm{V}]_{1}[\mathrm{~V}]_{2} \ldots[\mathrm{~V}]_{M}[\mathrm{CLASS}]$ and feeding to the frozen Text Encoder. The $[\mathrm{V}]_{m}(m \in\{1, \ldots, M\})$ is the vectors with the same dimension as word embeddings, and $M$ is a hyperparameter specifying the number of context tokens. The prediction probability is further computed as follows:
\begin{equation}
	p(y=i \mid \boldsymbol{x})=\frac{\exp \left(\cos \left(\mathcal{T}\left(\boldsymbol{t}_{i}\right), \boldsymbol{f}\right) / \tau\right)}{\sum_{j=1}^{K} \exp \left(\cos \left(\mathcal{T}\left(\boldsymbol{t}_{j}\right), \boldsymbol{f}\right) / \tau\right)}.
\end{equation}\\
\textbf{Visual Prompt Tuning (VPT)} 
aims to fine-tune the pre-trained models with only a few learnable parameters. Given the input image tokens $\boldsymbol{x}_i, i \in {1, ..., l}$ with length $l$, the trainable vectors $\boldsymbol{P}=\left\{\boldsymbol{p}^{k} \in \mathbb{R}^{d} \mid k \in \mathbb{N}, 1 \leq k \leq p\right\}$, where $d$ is the dimension of the patch embeddings and $p$ is the number of prompts, are cat with the original tokens and fed into the frozen encoders $Input = cat[\boldsymbol{x} \cdot \boldsymbol{P}]$ in certain orders.

\subsection{Overview of HiPTune}
We proposed a HiPTune that adopts hierarchical prompts to find multiple classification criteria from different semantic spaces and further integrates them, aiming to form more robust classification criteria for each input sample. Shown in Fig.~\ref{fig_HiPTune}, first, we construct the Visual Prompt Tree (VP-Tree) to gain the various semantics spaces tailored for each attack detection task. Then, given the input image $\boldsymbol{x}$, we adaptively select the most suitable prompt from each level of the VP-Tree by our Adaptive Prompt Pruning (APP) module. In addition, we integrate the prompt candidates from each selected level of the APP to describe the complete features of the input sample and feed it to the text encoder for better alignment through our Dynamic Prompt Integration (DPI) mechanism.

\subsection{Visual Prompt Tree (VP-Tree)}
Inspired by the work of~\cite{jia2022visual, yu2024visual}, learnable prompt parameters can be used to adapt a pretrained encoder, steering its output features to better align with downstream tasks. In the context of the UAD task, the diversity and high realism of forgery techniques pose significant challenges. To address this, we propose a hierarchical prompt tree designed to deliver coarse-to-fine guidance throughout the layered processing of a Transformer. At each block, the prompts encourage the encoder to refine its feature representations progressively, improving discrimination for classification. Notably, our prompts are class-agnostic in the sense that they target the distinction between authentic and forged samples in general, applying a specific prompt to each “real–fake” sample pair to capture their distinguishing characteristics.

\textbf{Define the levels.} 
We can obtain a hierarchical relationship between each attack type based on the spoof and forgery types from Fig.~\ref{fig_HiPTune}. Apart from the subjective categorization from the dataset, the features of different attacks demonstrate a more compact grouping relative to other categories while maintaining distinct independence within each category. For a VP-Tree with $N$ degrees, the level ${Tree}^{i} (i \in {1, 2, ..., N})$ has $n$ prompts, and $n$ may vary according to the set-ups. In our case, we define $3$-levels for the VP-Tree shown as Fig.~\ref{fig_HiPTune} (a). The first level distributes the most obvious data as physical and digital attack subtrees. The physical subtree is first separated into 2D and 3D attacks, which represent the second level of the VP-Tree. Print, replay, and cutouts are three key types for the level 3 prompts of a 2D attack. As for the 3D attacks, the masks' material is always considered to be the spoof clues, so we define transparent, plaster, and resin prompts within the 3D attack prompt. Concerning the digital subtree, we categorize the digital attacks into $3$ kinds according to their forgery techniques. The digital manipulation method includes attribute editing, face swapping, and video-driven faces. Adversarial attacks are divided into pixel-level and semantic-level attacks for forgery areas. We also define $3$ sub-prompts for the digital generation attacks: ID-consistent generation, style transfer, and prompt-based editing forgeries.

It is worth noting that to ensure continuity in the properties and semantics at each level, each subsequent prompt partially inherits semantic information from its preceding prompt. Specifically, given a prompt defined in $\boldsymbol{p}^0 \in \mathbb{R}^{l\times D}$, we apply an inheritance coefficient $\alpha$ to generate the inherited prompt $\boldsymbol{p}^1=cat[\boldsymbol{p}^1 \in \mathbb{R}^{(l-\alpha l)\times D},\boldsymbol{p}^1 \in \mathbb{R}^{\alpha l\times D}]$. This inherited component is then proportionally compressed, every $\frac{\alpha }{1} \cdot l$ tokens have been calculated the mean value and concatenated to the next-level prompt. This design ensures that the new prompt retains essential characteristics from its parent while introducing its own fine-grained, task-specific distinctions.

\textbf{Initialize the prompts.} 
After we obtain the VP-Tree's structure, we define the prompts for each level with random initialization in the following form:
\begin{equation}
	\boldsymbol{p}=[\mathrm{V}]_{1}[\mathrm{~V}]_{2} \ldots[\mathrm{~V}]_{l},
\end{equation}
where each $[\mathrm{V}]_{l}(l \in\{1, \ldots, L\})$ is the learnable vectors with the length of $L$ at the same dimension with the patch embeddings (i.e., 768 for CLIP image encoder).

\textbf{Train and integrate the prompts.} 
In order to force each prompt to learn the classification criteria specific to each attack type, we train the prompts in a supervised manner during the first stage. Given the input $\boldsymbol{x} \in \mathbb{R}^{C \times H \times W}$, where $C, H, W$ represent the channel, height, and width of the input images, we manually record the attack types and select the corresponding prompts at each level accordingly. Following the approach of XX, we insert the prompt tokens between the original image tokens and the image class token. This design allows the prompts to effectively modulate the flow of information from the input representations to the classification token, encouraging the encoder to integrate the prompt’s guidance during feature aggregation and thereby refining task-relevant representations. The final input of each Transformer block is formulated as Eq.~\ref{eq_VP-Tree_prompts}:
\begin{equation}
	{\mathcal{I}}_{input} = cat[\boldsymbol{x}_{cls}, \boldsymbol{p}, (\boldsymbol{x}_1, ..., \boldsymbol{x}_l)]
	\label{eq_VP-Tree_prompts}
\end{equation}
where $(\boldsymbol{x}_{cls}, \boldsymbol{x}_1, ..., \boldsymbol{x}_l) \in \mathbb{R}^{l+1, dim}$ is the image tokens after the patch embeddings. $l$ represents the length of image tokens, and $dim=768$ is the embedding dimension. The $\boldsymbol{p}$ is the selected prompt for each level.

As illustrated in the figure, we integrate different levels of prompts into specific layers of the Transformer blocks. As the Transformer progressively deepens, the semantic representation of tokens becomes increasingly precise. To leverage this property, we insert the first-level prompt after $N_0$ blocks—once the features have developed sufficient basic representational capacity—focusing on coarse discrimination between physical and digital attacks. Subsequently, after an additional $N_1$ and $N_2$ blocks, we introduce the second and third-level prompts, which provide increasingly fine-grained semantic guidance. Finally, the features undergo further processing through $N_3$ additional blocks to yield the final representation for classification.

After fully training, the VP-Tree to represent the classification criteria for multiple semantic spaces, the Adaptive Prompt Pruning (APP) module will learn to select the most suitable prompts for each sample.

\begin{table*}[]
	\centering
	\caption{Results on UniAttackData+. CLIP-V~\cite{radford2021learning} indicates removing the text encoder from CLIP. CoOp~\cite{zhou2022learning} (unified/specific) represents the CoOp without/with the class-specific context.}
	\scalebox{1.12}{
		\begin{tabular}{lcccc|cccc|cccc}
			\hline
			\multirow{2}{*}{Model} & \multicolumn{4}{c|}{P1: ID}    & \multicolumn{4}{c|}{P2: cross-attack-methods} & \multicolumn{4}{c}{P3.: cross-attacks-types} \\
			& ACER  & AUC   & ACC    & EER   & ACER      & AUC       & ACC       & EER       & ACER      & AUC       & ACC       & EER      \\ \hline
			CLIP-V~\cite{radford2021learning}                 & 9.36  & 92.46 & 88.05  & 9.36  & 28.77     & 71.99     & 78.49     & 28.77     & 49.87     & 43.12     & 58.08     & 49.87    \\
			CLIP~\cite{radford2021learning}                   & 9.93  & 95.22 & 83.75  & 9.93  & 22.89     & 72.79     & 76.15     & 22.89     & 34.37     & 58.09     & 59.90      & 34.37    \\
			CoOp~\cite{zhou2022learning}          & 14.76 & 90.89 & 83.92  & 14.76 & 43.14     & 59.64     & 47.83     & 43.14     & 72.00        & 20.89     & 27.18     & 72.00       \\
			CDCN++~\cite{yu2020searching}                 & 12.77 & 94.53 & 90.89  & 11.72 & 15.23     & 91.78     & 84.72     & 15.22     & 37.55     & 67.90      & 58.67     & 38.46    \\
			CFPL-FAS~\cite{liu2024cfpl}               & 13.93 & 93.77 & 96.24 & 11.65 & 16.10      & 86.34     & 83.09     & 16.10      & 29.60      & 73.50      & 71.46     & 29.60     \\ \hline
			MoE-FFD~\cite{kong2024moe}                & 17.30  & 91.90  & \textbf{96.50}   & 15.69 & 30.53     & 70.09     & \textbf{92.82}     & 31.39     & 46.01     & 51.27     & \textbf{89.14}     & 47.97    \\
			LLA-Net~\cite{nguyen2024laa}                & 10.57 & 95.64 & 95.97  &  \textbf{7.61}  & 18.34     & 82.97     & 89.86     & 21.31     & 36.58     & 57.02     & 68.25     & 42.20     \\ \hline
			FA$^{3}$-CLIP~\cite{li2025fa}                & 8.26  & 93.96 & 94.53  & 8.45  & 35.81     & 63.89     & 88.51     & 47.37     & 44.44     & 48.34     & 56.95     & 41.05    \\
			\rowcolor{gray!30}
			\textbf{HiPTune}                & \textbf{7.31}  & \textbf{97.87} & 90.02  & \textbf{7.61}  & \textbf{14.26}     & \textbf{92.50}      & 90.26     & \textbf{14.71}     & \textbf{25.68}     & \textbf{77.62}     & 69.34     & \textbf{26.02}    \\ \hline
		\end{tabular}
	}
	\label{tab_UniAttackData+}
\end{table*}
\subsection{Adaptive Prompt Pruning (APP)}
As shown in Fig.~\ref{fig_HiPTune} (a) and (c), aiming to explore classification criteria from different semantic spaces, the APP module adopts a hierarchical pruning mechanism to select the most suitable prompt for each sample. Given the input image $\boldsymbol{x} \in \mathbb{R}^{C \times H \times W}$, the patch embedding first projects the image into the visual embedding dimension and reshapes $\boldsymbol{x}$ into a series of image tokens with length $L$: $\boldsymbol{x}:\mathbb{R}^{C \times H \times W} \rightarrow \mathbb{R}^{L \times D}$. In our case, $C=3$ represents the channels for the image, and $H, W$ are both $224$ for the input image, the token's length $L=196$, and the dimension of visual embedding is $D=768$. After adding the positional embeddings, a class token is linked with the image token, and then the image token $\boldsymbol{x} \in \mathbb{R}^{197 \times 768}$ is fed to the Transformer blocks. Through $N_0$ blocks, the image tokens and the level-1 prompts are fed into the APP.

To select the most appropriate prompt, we first designate the front-most element (the first token after concatenation, with length 1) as the representative of each prompt. For the input image tokens, we use their $[CLS]$ token as the representation of the entire image. We then compute the attention scores between all prompt representatives and the image $[CLS]$ token. After applying the softmax function to normalize these scores, we select the single most relevant prompt using a $TOP_{k=1}$ strategy. This selected prompt is then inserted back into the sequence of image tokens to guide further encoding. The overall procedure is formally described as follows:
\begin{equation}
    index = TOP_{k=1}[Attention(\textbf{p}[0,:]\cdot x[0,:,:])],
\end{equation}
where $Attention(\cdot)$ represents the attention calculation, and the $index$ is the selection index of the most suitable prompt.

It is important to note that to ensure each prompt accurately captures its intended semantic information during selection, we supervise the training of prompts using ground-truth labels at each level for different samples in the training set. This enables each prompt to specialize in discriminating its designated category or property. Once trained, the resulting VP-Tree is integrated into the model as a fixed module, supporting the APP (Adaptive Prompt Pooling) selection mechanism during downstream training.

\subsection{Dynamic Prompt Interaction (DPI)}
CLIP's default templates, ~\textit{e.g.},~``\texttt{This is a photo of [CLASS]}'', fail to describe the detailed semantic information for the multiple attack types, which caused the performance drops encountering advanced attacks. Moreover, the unified prompts in the CoOp still can not represent the differences between various attacks and the live faces. However, as mentioned above, our VP-Tree constructs a hierarchical prompt tree for the different classification criteria, and our APP obtains the probabilities of the suitable candidate prompts. Thus, we proposed a Dynamic Prompt Interaction (DPI) to project the visual prompts to the text encoder to fully describe the abundant features corresponding to each category.

Specifically, shown as Fig.~\ref{fig_HiPTune} (b), after the APP module selects the appropriate prompts at each hierarchical level, we stack these prompts in order to form a combined sequence. This stacked sequence is then processed using self-attention, allowing information exchange across levels. Following the softmax operation, we compute a weighted sum of the three hierarchical prompts to produce the final unified prompt. Finally, this prompt is projected from the original image feature dimension to the target text feature dimension through a dedicated mapping layer.


\section{Experiments}
\subsection{Experiment Settings}
{\noindent \textbf{Datasets, Protocols, and Evaluation Metrics.}} We evaluate the effectiveness of the proposed UniAttackData+ and HiPTune using three existing UAD datasets. For the \textit{UniAttackData+} dataset, we conduct experiments on the three proposed protocols. For \textit{JFSFDB}~\cite{yu2024benchmarking} dataset, we only conduct cross-domain experiments on both separate training on each single task (\ie, PAD or FFD) and joint training within both two tasks (\ie, UAD). Specifically, separate training is conducted by training two distinct models on datasets {SiW, 3DMAD, HKBU} for PAD, and {FF++} for FFD, respectively. The trained models are then evaluated on cross-domain datasets {MSU, 3DMask, ROSE} for PAD and {DFDC, CelebDFv2} for FFD. Instead of training separate models, joint training involves learning a unified model for both tasks by leveraging all data from {SiW, 3DMAD, HKBU, FF++}. The unified model is then evaluated on cross-domain datasets {MSU, 3DMask, ROSE, DFDC, CelebDFv2} for UAD. For \textit{UniAttackData}~\cite{fang2024unified} dataset: (1) Protocol 1 evaluates performance under a unified attack detection task with all attack types included in training and testing; (2) Protocol 2 assesses generalization to unseen attack types using a leave-one-type-out strategy, with Protocols 2.1 and 2.2 focusing on unseen physical and digital attacks, respectively. Following the setup in FA$^{3}$-CLIP~\cite{li2025fa}, we also conduct experiments under Protocol 1.1, 1.2, and 1.3. Protocols 1.1 and 1.2 independently evaluate the digital category by excluding deepfake and adversarial attacks, respectively, from training and validation, testing them only on separate identity sets. Protocol 1.3 includes all digital subtypes with standard distribution. These protocols assess the model’s robustness in unified face attack detection across both physical and digital threats. In addition, we construct a domain generalization-based UAD benchmark using the JFSFDB, UniAttackData, and UniAttackData+ datasets, where any two are combined for training and the remaining one is used for testing. 

We evaluate performance using four metrics: (1) ACER, the average of the false rejection rate (FRR) and the false acceptance rate (FAR), (2) AUC, reflecting the model’s theoretical performance, (3) EER, representing the error rate at the optimal threshold, and (4) ACC, indicating the overall classification accuracy.

{\noindent \textbf{Implementation Details.}} We use a $3$-level VP-Tree with a prompt length of $40$ and a frozen CLIP model with ViT/b-16 backbone, classifying ``live" and ``fake". The initial learning rate is $10^{-5}$ with the Adam optimizer. Images are resized to $224 \times 224$ with random cropping and flipping. The initialization process of VP-Tree has been trained for $100$ epochs (batch size: $32$), and followed by full model training for $500$ epochs (batch size: $32$). The image tokens are first fed into $N_0=3$ blocks to gain the initial representation abilities. In our case, $N_1=N_2=N_3=3$ blocks are followed by each selection of prompts.

\begin{table*}[t]
	\centering
	\caption{Results on the JFSFDB with three cross-domain tasks: PAD, FFD, and UAD. For PAD, the training set consists of SiW, 3DMAD, and HKBU, while the testing set includes MSU, 3DMask, and ROSE. For FFD, the training set is FF++, and the testing set comprises DFDC, CelebDFv2. For UAD, the training set combines SiW, 3DMAD, HKBU, and FF++, and the testing set includes MSU, 3DMask, ROSE, DFDC, and CelebDFv2. ↓/↑ indicates that smaller/larger values correspond to better performance. The Avg. represents the mean result for all testing sets.
	}
	\scalebox{0.95}{
		\begin{tabular}{|l|l|ccc|cc|ccccc|c|}
			\hline
			\multirow{2}{*}{Metrics}  & \multirow{2}{*}{Method} & \multicolumn{3}{c|}{PAD}                                                                   & \multicolumn{2}{c|}{FFD}                             & \multicolumn{5}{c|}{UAD}                                                                                                                                               & \multirow{2}{*}{Avg.} \\ \cline{3-12}
			&                         & \multicolumn{1}{c|}{MSU}            & \multicolumn{1}{c|}{3DMask}         & ROSE           & \multicolumn{1}{c|}{DFDC}           & Celeb-DFv2     & \multicolumn{1}{c|}{MSU}            & \multicolumn{1}{c|}{3DMask}         & \multicolumn{1}{c|}{ROSE}           & \multicolumn{1}{c|}{DFDC}           & Celeb-DFv2     &                       \\ \hline
			\multirow{14}{*}{EER(\%)↓} 
			& MesoNet~\cite{afchar2018mesonet}                 & \multicolumn{1}{c|}{21.90}          & \multicolumn{1}{c|}{55.82}          & 36.81          & \multicolumn{1}{c|}{46.16}          & 38.78          & \multicolumn{1}{c|}{22.86}          & \multicolumn{1}{c|}{54.76}          & \multicolumn{1}{c|}{36.18}          & \multicolumn{1}{c|}{49.81}          & 46.94          & 41.00                 \\ \cline{2-13} 
			& Xception~\cite{chollet2017xception}                & \multicolumn{1}{c|}{18.57}          & \multicolumn{1}{c|}{42.30}          & 18.13          & \multicolumn{1}{c|}{38.92}          & 24.07          & \multicolumn{1}{c|}{17.14}          & \multicolumn{1}{c|}{22.21}          & \multicolumn{1}{c|}{28.82}          & \multicolumn{1}{c|}{40.75}          & 29.16          & 28.00                 \\ \cline{2-13} 
			& CDCN++~\cite{yu2020searching}                  & \multicolumn{1}{c|}{22.80}          & \multicolumn{1}{c|}{52.76}          & 32.01          & \multicolumn{1}{c|}{44.75}          & 28.19          & \multicolumn{1}{c|}{31.90}          & \multicolumn{1}{c|}{43.83}          & \multicolumn{1}{c|}{33.19}          & \multicolumn{1}{c|}{44.58}          & 29.68          & 36.37                 \\ \cline{2-13} 
			& DeepPixel~\cite{george2019deep}               & \multicolumn{1}{c|}{15.71}          & \multicolumn{1}{c|}{46.42}          & 28.23          & \multicolumn{1}{c|}{36.82}          & \textbf{22.81} & \multicolumn{1}{c|}{13.33}          & \multicolumn{1}{c|}{40.19}          & \multicolumn{1}{c|}{32.80}          & \multicolumn{1}{c|}{37.38}          & 19.48          & 29.32                 \\ \cline{2-13} 
			& JFSFDB-Uni~\cite{yu2024benchmarking}              & \multicolumn{1}{c|}{-}              & \multicolumn{1}{c|}{-}              & -              & \multicolumn{1}{c|}{-}              & -              & \multicolumn{1}{c|}{6.67}           & \multicolumn{1}{c|}{43.60}          & \multicolumn{1}{c|}{11.64}          & \multicolumn{1}{c|}{30.60}          & 18.80          & 22.26                 \\ \cline{2-13} 
			& CLIP~\cite{radford2021learning}                    & \multicolumn{1}{c|}{12.86}          & \multicolumn{1}{c|}{26.74}          & 14.61          & \multicolumn{1}{c|}{23.14}          & 27.16          & \multicolumn{1}{c|}{10.24}          & \multicolumn{1}{c|}{23.58}          & \multicolumn{1}{c|}{26.86}          & \multicolumn{1}{c|}{\textbf{23.76}} & 27.34          & 21.63                 \\ \cline{2-13} 
			& CoOp~\cite{zhou2022learning}           & \multicolumn{1}{c|}{25.95}          & \multicolumn{1}{c|}{17.02}          & 12.06          & \multicolumn{1}{c|}{39.42}          & 41.21          & \multicolumn{1}{c|}{28.57}          & \multicolumn{1}{c|}{17.77}          & \multicolumn{1}{c|}{14.39}          & \multicolumn{1}{c|}{36.69}          & 39.84          & 27.29                 \\ \cline{2-13} 
			& FA$^{3}$-CLIP~\cite{li2025fa}               & \multicolumn{1}{c|}{6.90}           & \multicolumn{1}{c|}{18.87}          & \textbf{9.61}  & \multicolumn{1}{c|}{22.46}          & 25.51          & \multicolumn{1}{c|}{4.29}           & \multicolumn{1}{c|}{24.00}          & \multicolumn{1}{c|}{10.77}          & \multicolumn{1}{c|}{24.53}          & 30.02          & 17.70                 \\ \cline{2-13} 
			\rowcolor{gray!30}
			& HipTune                 & \multicolumn{1}{c|}{\textbf{6.56}}  & \multicolumn{1}{c|}{\textbf{11.57}} & 12.82          & \multicolumn{1}{c|}{\textbf{20.86}} & 28.77          & \multicolumn{1}{c|}{\textbf{3.84}}  & \multicolumn{1}{c|}{\textbf{11.72}} & \multicolumn{1}{c|}{\textbf{8.99}}  & \multicolumn{1}{c|}{24.92}          & \textbf{12.21} & \textbf{14.23}        \\ \hline
			\multirow{14}{*}{AUC(\%)↑} 
			& MesoNet~\cite{afchar2018mesonet}                 & \multicolumn{1}{c|}{85.33}          & \multicolumn{1}{c|}{44.54}          & 68.05          & \multicolumn{1}{c|}{54.72}          & 65.10          & \multicolumn{1}{c|}{84.01}          & \multicolumn{1}{c|}{45.87}          & \multicolumn{1}{c|}{70.00}          & \multicolumn{1}{c|}{50.57}          & 55.03          & 62.32                 \\	\cline{2-13}
			& Xception~\cite{chollet2017xception}                & \multicolumn{1}{c|}{90.62}          & \multicolumn{1}{c|}{63.88}          & 90.15          & \multicolumn{1}{c|}{64.86}          & 84.27          & \multicolumn{1}{c|}{91.56}          & \multicolumn{1}{c|}{75.49}          & \multicolumn{1}{c|}{79.38}          & \multicolumn{1}{c|}{63.69}          & 78.87          & 78.28                 \\ \cline{2-13} 
			& CDCN++~\cite{yu2020searching}                 & \multicolumn{1}{c|}{82.65}          & \multicolumn{1}{c|}{47.66}          & 76.76          & \multicolumn{1}{c|}{56.65}          & 78.34          & \multicolumn{1}{c|}{78.93}          & \multicolumn{1}{c|}{61.54}          & \multicolumn{1}{c|}{76.39}          & \multicolumn{1}{c|}{56.42}          & 76.92          & 69.23                 \\ \cline{2-13} 
			& DeepPixel~\cite{george2019deep}               & \multicolumn{1}{c|}{95.17}          & \multicolumn{1}{c|}{56.87}          & 80.60          & \multicolumn{1}{c|}{67.54}          & 85.51          & \multicolumn{1}{c|}{93.97}          & \multicolumn{1}{c|}{66.65}          & \multicolumn{1}{c|}{74.58}          & \multicolumn{1}{c|}{66.36}          & 88.45          & 77.57                 \\ \cline{2-13} 
			& JFSFDB-Uni~\cite{yu2024benchmarking}              & \multicolumn{1}{c|}{-}              & \multicolumn{1}{c|}{-}              & -              & \multicolumn{1}{c|}{-}              & -              & \multicolumn{1}{c|}{97.99}          & \multicolumn{1}{c|}{67.10}          & \multicolumn{1}{c|}{95.57}          & \multicolumn{1}{c|}{75.90}          & 89.76          & 85.26                 \\ \cline{2-13} 
			& CLIP~\cite{radford2021learning}                    & \multicolumn{1}{c|}{95.82}          & \multicolumn{1}{c|}{80.40}          & 92.89          & \multicolumn{1}{c|}{86.08}          & 79.40          & \multicolumn{1}{c|}{96.39}          & \multicolumn{1}{c|}{86.13}          & \multicolumn{1}{c|}{81.07}          & \multicolumn{1}{c|}{84.06}          & 78.99          & 86.12                 \\ \cline{2-13} 
			& CoOp~\cite{zhou2022learning}           & \multicolumn{1}{c|}{83.73}          & \multicolumn{1}{c|}{91.94}          & 94.63          & \multicolumn{1}{c|}{64.46}          & 62.05          & \multicolumn{1}{c|}{81.58}          & \multicolumn{1}{c|}{89.27}          & \multicolumn{1}{c|}{93.62}          & \multicolumn{1}{c|}{68.56}          & 64.88          & 79.47                 \\ \cline{2-13} 
			& FA$^{3}$-CLIP~\cite{li2025fa}                & \multicolumn{1}{c|}{97.96}          & \multicolumn{1}{c|}{89.99}          & 95.52          & \multicolumn{1}{c|}{85.60}          & 83.15          & \multicolumn{1}{c|}{98.12}          & \multicolumn{1}{c|}{81.84}          & \multicolumn{1}{c|}{95.58}          & \multicolumn{1}{c|}{82.95}          & 74.80          & 88.55                 \\ \cline{2-13} 
			\rowcolor{gray!30}
			& HipTune                 & \multicolumn{1}{c|}{\textbf{98.62}} & \multicolumn{1}{c|}{\textbf{95.87}} & \textbf{98.62} & \multicolumn{1}{c|}{\textbf{89.72}} & \textbf{89.63} & \multicolumn{1}{c|}{\textbf{99.62}} & \multicolumn{1}{c|}{\textbf{95.43}} & \multicolumn{1}{c|}{\textbf{97.82}} & \multicolumn{1}{c|}{\textbf{90.88}} & \textbf{93.35} & \textbf{94.96}        \\ \hline
		\end{tabular}
	}
	\label{tab_JFSFDB}
\end{table*}

\subsection{Results on UniAttackData+}
We employ a range of CLIP-based baselines, including CLIP~\cite{radford2021learning}, CLIP-V~\cite{radford2021learning}, and CoOp~\cite{zhou2022learning} to evaluate the performance in UAD on our dataset. In addition, we assess algorithms specifically designed for PAD/FFD, such as CFPL-FAS~\cite{liu2024cfpl}, MoE-FFD~\cite{kong2024moe}, and LLA-Net~\cite{nguyen2024laa}, to examine their generalization capabilities. We also compare our approach with the recently released FA$^3$-CLIP, an algorithm tailored for unified attack detection. The detailed experimental results are presented in Tab.~\ref{tab_UniAttackData+}.

Protocol 1 is designed to evaluate the model’s generalization of face identity in the UAD task. As shown in Tab.~\ref{tab_UniAttackData+}, we can observe that CLIP-V~\cite{radford2021learning} achieves decent performance, with an ACER of 9.36\%. However, introducing fixed text templates or single-semantic learnable vectors does not bring performance benefits, such as the ACER of CLIP~\cite{radford2021learning} and CoOp~\cite{zhou2022learning} being 9.93\% and 14.76\%, respectively. These experimental results indicate that in the UAD task, naive category prompts cannot accurately describe highly challenging multi-attack types, resulting in ineffective differentiation from real faces. Although the recently released CDCN++\cite{yu2020searching}, CFPL-FAS\cite{liu2024cfpl}, MoE-FFD~\cite{kong2024moe}, and LLA-Net~\cite{nguyen2024laa} have achieved superior performance in PAD and FFD tasks, their effectiveness drops significantly on our UniAttackData+ dataset, with ACERs of 12.77\%, 13.93\%, 17.30\%, and 10.57\%, respectively. This suggests that these algorithms exhibit poor generalization across different attack categories, as models trained on PAD datasets fail to adapt effectively to FFD scenarios. FA$^{3}$-CLIP~\cite{li2025fa} solves UAD by combining spatial and frequency features in a dual-stream architecture. By employing an attack-agnostic prompt learning and bridging the gap between different attack types, it significantly improves UAD performance, such as ACER is 8.26\%. Our proposed HiPTune employs a visual prompt-based strategy that effectively guides the encoder toward informative features while preserving model efficiency. Furthermore, the hierarchical prompt tree captures multi-level, fine-grained classification criteria, enabling highly accurate and generalizable performance across diverse attack scenarios, with an ACER of just 7.31\%.

Protocol 2 evaluates the generalization of the model across attack methods in the UAD task, which is widely applicable in real-world scenarios. When encountering unseen attacks, algorithms that either lack text prompts or rely solely on simple prompts exhibit significantly degraded performance. For example, CLIP-V~\cite{radford2021learning} and CoOp~\cite{zhou2022learning} achieve ACERs of 28.77\% and 43.14\%, respectively. Although FA$^{3}$-CLIP~\cite{li2025fa} demonstrates promising results on Protocol 1, its effectiveness significantly declines when encountering unknown attack types, with an ACER of 35.81\%. This suggests that single-semantic prompt learning struggles to generalize across diverse attack categories. Unlike conventional prompt-matching methods, our prompts function solely as guidance rather than decision-making anchors, thereby reducing reliance on prior knowledge. This design enhances the model's generalization to unseen attack methods while ensuring robustness and interpretability, achieving an ACER of 14.26\%.

Protocol 3 is designed to evaluate whether models trained on simple attack types can retain high robustness when tested against complex and sophisticated attacks. We can observe that almost all methods have degraded performance, even to the point of random guessing, such as the ACER of CLIP-V~\cite{radford2021learning}, CoOp~\cite{zhou2022learning}, MoE-FFD~\cite{kong2024moe}, and FA$^{3}$-CLIP~\cite{li2025fa} exceeding 40\%. In contrast, our HiPTune model adaptively integrates diverse types of prompts and demonstrates superior performance. Despite being trained on limited samples, HiPTune leverages high-level prompts and auxiliary information from other attack types to support inference. This design enables the model to maintain strong generalization capabilities, even against advanced attacks, achieving an ACER of 25.68\%.

\begin{table*}[t]
	\centering
	\caption{Results on UniAttackData with standard Protocol 1, 1.1, 1.2, 1.3, and 2. ↓/↑ indicate that smaller/larger values correspond to better performance. The Avg. represents the mean ACER for Protocol 1, 1.1, 1.2, 1.3.}
	\scalebox{1.05}{
		\begin{tabular}{|l|cc|cc|cc|cc|c|cc|}
			\hline
			\multirow{2}{*}{Methods} & \multicolumn{2}{c|}{P1}           & \multicolumn{2}{c|}{P1.1}          & \multicolumn{2}{c|}{P1.2}          & \multicolumn{2}{c|}{P1.3}         & \multirow{2}{*}{Avg.} & \multicolumn{2}{c|}{P2}            \\ \cline{2-9} \cline{11-12} 
			& \multicolumn{1}{c|}{ACER(\%)↓} & AUC(\%)↑   & \multicolumn{1}{c|}{ACER}  & AUC   & \multicolumn{1}{c|}{ACER}  & AUC   & \multicolumn{1}{c|}{ACER} & AUC   &                       & \multicolumn{1}{c|}{ACER}  & AUC   \\ \hline
			ResNet-50                & \multicolumn{1}{c|}{1.35} & 99.79 & \multicolumn{1}{c|}{\textbf{5.92}}  & 91.25 & \multicolumn{1}{c|}{25.90} & 84.35 & \multicolumn{1}{c|}{4.92} & 98.84 & 9.52                  & \multicolumn{1}{c|}{34.60} & 87.89 \\ \hline
			ViT-B/16                 & \multicolumn{1}{c|}{5.92} & 97.00 & \multicolumn{1}{c|}{13.53} & 95.99 & \multicolumn{1}{c|}{5.22}  & 99.36 & \multicolumn{1}{c|}{3.20} & 99.18 & 6.97                  & \multicolumn{1}{c|}{33.69} & 83.77 \\ \hline
			CDCN~\cite{yu2020searching}                     & \multicolumn{1}{c|}{1.40} & 99.52 & \multicolumn{1}{c|}{12.32} & 93.89 & \multicolumn{1}{c|}{16.34} & 93.34 & \multicolumn{1}{c|}{4.41} & 97.68 & 8.62                  & \multicolumn{1}{c|}{34.33} & 77.46 \\ \hline
			CLIP~\cite{radford2021learning}                     & \multicolumn{1}{c|}{1.02} & 99.47 & \multicolumn{1}{c|}{14.81} & 86.74 & \multicolumn{1}{c|}{5.36}  & 99.17 & \multicolumn{1}{c|}{2.45} & 97.92 & 5.91                  & \multicolumn{1}{c|}{15.27} & 92.46 \\ \hline
			UniAttackDetection~\cite{fang2024unified}       & \multicolumn{1}{c|}{0.52} & 99.96 & \multicolumn{1}{c|}{11.73} & \textbf{98.81} & \multicolumn{1}{c|}{1.70}  & \textbf{99.85} & \multicolumn{1}{c|}{4.67} & 99.13 & 4.66                  & \multicolumn{1}{c|}{22.42} & 91.97 \\ \hline
			FA$^{3}$-CLIP~\cite{li2025fa}                    
			& \multicolumn{1}{c|}{0.36}      & 99.75 
			& \multicolumn{1}{c|}{9.57}      & 97.78                                    
			& \multicolumn{1}{c|}{1.43}  & \textbf{99.85} 
			& \multicolumn{1}{c|}{2.30} & \textbf{99.19} 
			& 3.42                  
			& \multicolumn{1}{c|}{18.81}     & 88.59     \\ \hline
			\rowcolor{gray!30}
			HiPTune  
			& \multicolumn{1}{c|}{\textbf{0.33}} & \textbf{99.99} 
			& \multicolumn{1}{c|}{7.63}      & 90.52       
			& \multicolumn{1}{c|}{\textbf{1.26}}      &  98.88      
			& \multicolumn{1}{c|}{\textbf{1.08}}     &  97.18     
			& \textbf{2.58}  & \multicolumn{1}{c|}{\textbf{10.38}} & \textbf{97.82} \\ \hline
		\end{tabular}
	}
	\label{tab_uniattackdata}
\end{table*}

\subsection{Results on JFSFDB Dataset}
The Tab.~\ref{tab_JFSFDB} presents the evaluation results of various methods across three tasks: PAD, FFD, and UAD. Two metrics are used for performance comparison: Equal Error Rate (EER\%) and Area Under Curve (AUC\%). The experiment compares traditional methods (\eg, MesoNet~\cite{afchar2018mesonet}, Xception~\cite{chollet2017xception}, DeepPixel~\cite{george2019deep}, CDCN++~\cite{yu2020searching}, JFSFDB-Uni~\cite{yu2024benchmarking}), CLIP-based baselines (CLIP~\cite{radford2021learning}, CoOp~\cite{zhou2022learning}, FA$^{3}$-CLIP~\cite{li2025fa}), and the proposed method HiPTune.

In tasks PAD and FFD, our HiPTune achieves the best results in the AUC metric, such as 98.62\%, 95.87\%, and 98.62\% on datasets MSU, 3DMask, and ROSE, and 89.72\% and 89.63\% on datasets DFDC and Celeb-DFv2, respectively. However, in terms of EER, HiPTune shows slightly weaker generalization on the ROSE and Celeb-DFv2 datasets. We attribute this to a class imbalance issue, where the model tends to learn better representations for the majority class (\eg, real faces). As a result, it achieves strong overall ranking performance (reflected in a high AUC). However, during testing, the decision threshold cannot effectively balance the trade-off between the two classes, leading to suboptimal EER performance. After training the UAD model with joint physics and digital forgery data, our HiPTune further improved the performance of AUC on most cross-dataset test results, such as MSU improving by 1.00\% (98.62\%→99.62\%), DFDC improving by 1.16\% (89.72\%→90.88\%), and Celeb-DFv2 improving by 3.72\% (89.63\%→93.35\%). These experimental results indicate that our HiPTune can combine the diversity and complementarity of PAs and DFs to improve its performance in PAD and FFD tasks, respectively. However, for datasets 3DMask and ROSE, the performance of UAD slightly decreased, such as 95.87\% decreasing to 95.43\%, 98.62\% decreasing to 97.82\%, respectively. This indicates that introducing the FF++ dataset into the training set can impair the model's ability to recognize 3D masks. We analyze that in our HiPTune framework, prompts are organized hierarchically by attack type. If a large influx of digital attacks (like from FF++) dominates the prompt learning phase, it could bias the learned prompts or pruning decisions, weakening the semantic alignment for physical branches (like 3D masks).

\subsection{Results on UniAttackData}
Tab.~\ref{tab_uniattackdata} presents the performance of various methods on the UniAttackData under standard Protocols 1, 1.1, 1.2, 1.3, and 2, where lower ACER and higher AUC indicate better results. The Avg. column summarizes the average ACER across Protocols 1, 1.1, 1.2, and 1.3. HiPTune achieves the best overall performance, with the lowest average ACER (2.58\%) and the highest AUC (99.99\%) under Protocol 1, as well as strong performance in Protocol 2 (ACER: 10.38\%, AUC: 97.82\%). On Protocol 1, 1.2, and 1.3, HiPTune achieves the best ACER values of 0.33\%, 1.26\%, and 1.08\%, respectively, and obtains the best mean ACER (2.58\%). It indicates that HiPTune has the best generalization for facial identity information, such as the strict identity partitioning in the training, validation, and testing sets of Protocol 1.3. When physical presentation attacks and adversarial attacks are used as training sets, HiPTune has poorer generalization to deepfake attacks than ResNet-50, such as 7.63\% v.s. 5.92\% for ACER. On the contrary, the performance of ResNet-50 has significantly decreased, such as achieving an ACER value of 25.90\% on Protocol 1.2. However, HiPTune still maintains its leading performance, achieving an ACER of 1.26\%. This proves the effectiveness of using hierarchical prompt learning in HiPTune, where the model uses a hierarchical routing mechanism to select the most relevant prompts for each input image. 

Finally, on the cross-attack category Protocol 2, where the model is trained on physical presentation attacks and tested on digital forgery (including opposite training and testing), our HiPTune demonstrates an absolute advantage, achieving an ACER of 10.38\% and an AUC of 97.82\%. HiPTune demonstrates strong generalization across different attack categories, due to the following points: (1) Hierarchical prompt modeling captures shared and distinct semantics. VP-Tree organizes attacks hierarchically by semantic similarity. High-level prompts (``physical" \vs ``digital") provide coarse but generalizable decision boundaries, while low-level prompts capture fine-grained modality-specific traits. This structure allows the model to transfer knowledge from one modality (\eg, physical) to another (\eg, digital), by reusing the shared higher-level decision boundaries. (2) Dynamic prompt selection tailored to each sample. The APP module ensures that, at inference time, the model can select the most appropriate semantic pathway, even if that pathway wasn't explicitly optimized during training. This enables robust out-of-distribution generalization when tested on unseen attack types.

\subsection{Results on cross-domain UAD benchmark}
\begin{table}[!t]
	\centering
	\caption{Results on cross-domain UAD benchmark.}
	\scalebox{1.04}{
		\begin{tabular}{|l|cc|cc|cc|}
			\hline
			\multirow{2}{*}{Method} & \multicolumn{2}{c|}{U\&U+→J}       & \multicolumn{2}{c|}{J\&U+→U}       & \multicolumn{2}{c|}{J\&U→U+}       \\ \cline{2-7} 
			& \multicolumn{1}{c|}{ACER}  & AUC   & \multicolumn{1}{c|}{ACER}  & AUC   & \multicolumn{1}{c|}{ACER}  & AUC   \\ \hline
			CLIP-V                  & \multicolumn{1}{c|}{45.12} & 56.26 & \multicolumn{1}{c|}{13.15} & 91.86 & \multicolumn{1}{c|}{39.81} & 63.4  \\ \hline
			CLIP                    & \multicolumn{1}{c|}{44.42} & 58.10  & \multicolumn{1}{c|}{16.59} & 90.14 & \multicolumn{1}{c|}{39.36} & 64.08 \\ \hline
			CoOp                    & \multicolumn{1}{c|}{38.21} & 65.84 & \multicolumn{1}{c|}{30.28} & 76.11 & \multicolumn{1}{c|}{29.77} & 76.77 \\ \hline
			\rowcolor{gray!30}
			HiPTune                 & \multicolumn{1}{c|}{\textbf{32.93}} & \textbf{70.89} & \multicolumn{1}{c|}{\textbf{8.55}}  & \textbf{94.67} & \multicolumn{1}{c|}{\textbf{25.93}} & \textbf{83.05} \\ \hline
		\end{tabular}
	}
	\label{tab_cross_domain_UAD}
\end{table}

Tab.~\ref{tab_cross_domain_UAD} presents the results of various methods, \ie, CLIP-V~\cite{radford2021learning}, CLIP~\cite{radford2021learning}, CoOp~\cite{zhou2022learning}, and HiPTune on the cross-domain UAD benchmark, designed to evaluate the generalization ability of the models under domain shifts. The benchmark consists of three protocols, each involving training and testing across different combinations of datasets: UniAttackData (U), UniAttackData+ (U+), and JFSFDB (J). Among the compared methods, HiPTune achieves the best performance across all three cross-domain settings. For example, when the test set is J, U, and U+, our HiPTune achieves ACER values of 32.93\%, 8.55\%, and 25.93\%, respectively, which are significantly better than the compared algorithms. The superiority of domain generalization in our analysis comes from the following two points: (1) HiPTune constructs a VP-Tree that models attack types in a coarse-to-fine hierarchy, enabling the model to distinguish diverse attack semantics across different domains. This structured representation ensures that even if an unseen domain presents a new variant of an attack, the model can still leverage higher-level prompts to make correct predictions. (2) The APP module dynamically selects the most appropriate prompts for each sample based on its visual features, rather than relying on fixed global parameters. This flexibility allows the model to adapt to distribution shifts without retraining, thereby improving generalization to unseen domains.

\subsection{Ablation Studies}
\begin{figure}[t]
	\centering
	\includegraphics[width=0.99\linewidth]{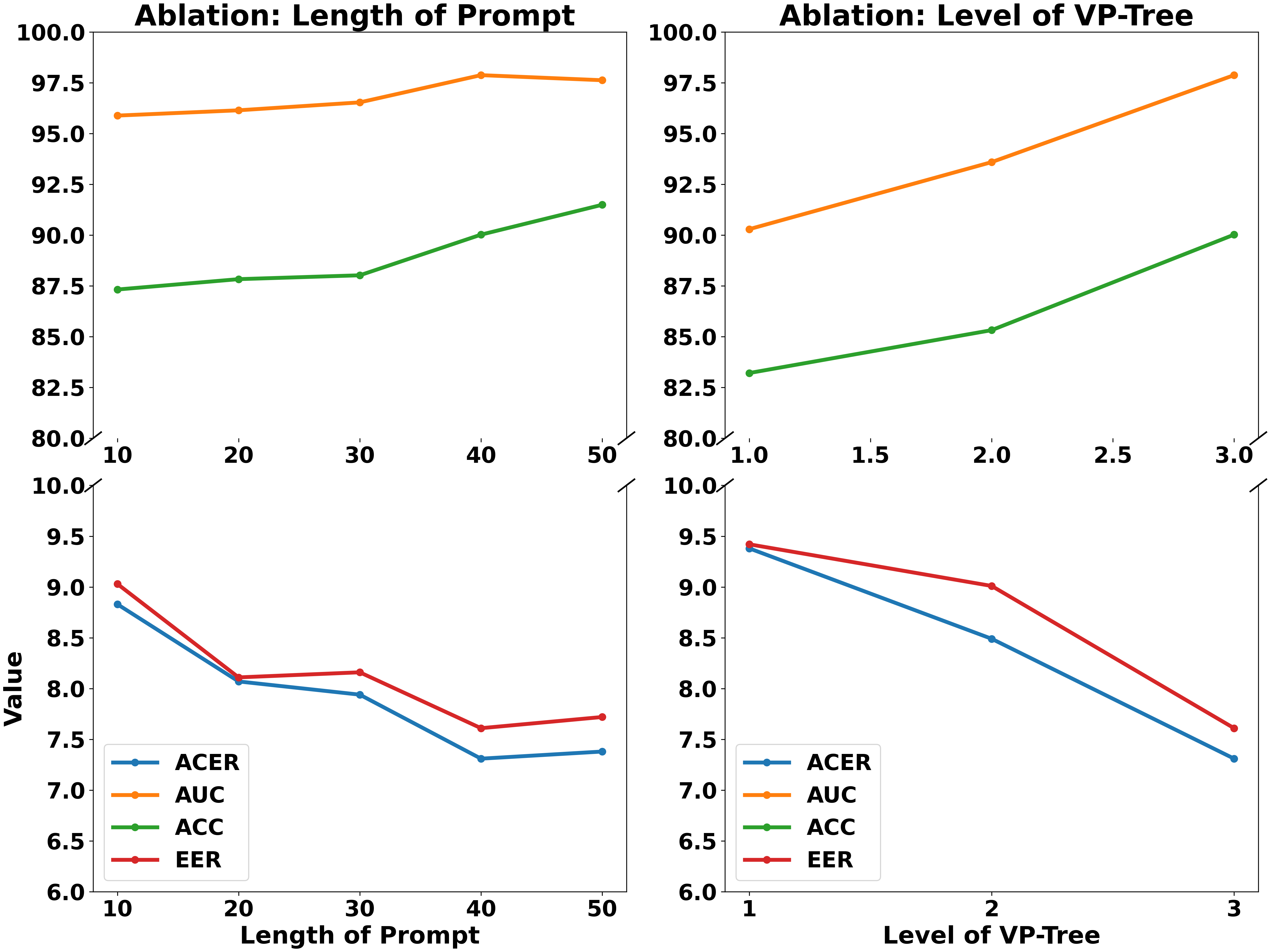}
	\caption{
		Ablation studies on the level of VP-Tree and the length of the prompt.
	}
	\label{fig:Ablation_Studies}
\end{figure}

\textbf{Effect of VP-Tree Depth.}
To evaluate the contribution of the hierarchical structure in VP-Tree, we conduct experiments on protocol 1 of UniAttackData+ by varying the depth of the tree from 1 to 3. As shown in Fig.~\ref{fig:Ablation_Studies}, deeper trees consistently improve the performance across all metrics. Specifically, the full three-level VP-Tree achieves the best ACER (7.31\%) and EER (7.61\%), indicating that more fine-grained semantic grouping leads to more robust classification criteria. This demonstrates the benefit of modeling attack diversity hierarchically.

\textbf{Effect of Prompt Length.}
We also study the impact of the prompt length in our framework. As presented in Fig.~\ref{fig:Ablation_Studies}, increasing the length from 10 to 40 improves performance steadily. The model achieves optimal results at a length of 40, with an ACER of 7.31\% and an AUC of 97.87\%. Further increasing the length to 50 offers marginal gains in accuracy but slightly degrades ACER and EER, suggesting a trade-off between prompt expressiveness and overfitting. Thus, we adopt 40 as the default length in our experiments.

\section{Conclusion}
\label{sec:Conclusion}
In this paper, we first present a large-scale Unified Face Attack Benchmark (HiFAB), which is the most extensive and comprehensive UAD dataset to date. It includes $14$ kinds of physical attacks, $40$ digital attacks, and $697,347$ videos in total. Every attack is applied to $2,875$ identities to reduce the ID noise. Furthermore, we introduce a novel Hierarchical Prompt Tuning framework (HiPTune) that adaptively integrates classification criteria from various semantic spaces, allowing tailored classification criteria for each sample. Finally, the proposed HiPTune demonstrates superior performance and generalization across three unified face forgery detection benchmarks.

\ifCLASSOPTIONcaptionsoff
  \newpage
\fi

%
\bibliographystyle{IEEEtran}
\bibliography{IEEEabrv,main}

\begin{thebibliography}{10}
\providecommand{\url}[1]{#1}
\csname url@samestyle\endcsname
\providecommand{\newblock}{\relax}
\providecommand{\bibinfo}[2]{#2}
\providecommand{\BIBentrySTDinterwordspacing}{\spaceskip=0pt\relax}
\providecommand{\BIBentryALTinterwordstretchfactor}{4}
\providecommand{\BIBentryALTinterwordspacing}{\spaceskip=\fontdimen2\font plus
\BIBentryALTinterwordstretchfactor\fontdimen3\font minus \fontdimen4\font\relax}
\providecommand{\BIBforeignlanguage}[2]{{%
\expandafter\ifx\csname l@#1\endcsname\relax
\typeout{** WARNING: IEEEtran.bst: No hyphenation pattern has been}%
\typeout{** loaded for the language `#1'. Using the pattern for}%
\typeout{** the default language instead.}%
\else
\language=\csname l@#1\endcsname
\fi
#2}}
\providecommand{\BIBdecl}{\relax}
\BIBdecl

\bibitem{zhang2020casia}
S.~Zhang, A.~Liu, J.~Wan, Y.~Liang, G.~Guo, S.~Escalera, H.~J. Escalante, and S.~Z. Li, ``Casia-surf: A large-scale multi-modal benchmark for face anti-spoofing,'' \emph{IEEE Transactions on Biometrics, Behavior, and Identity Science}, vol.~2, no.~2, pp. 182--193, 2020.

\bibitem{Chingovska_BIOSIG-2012}
I.~Chingovska, A.~Anjos, and S.~Marcel, ``On the effectiveness of local binary patterns in face anti-spoofing,'' in \emph{BIOSIG}, 2012.

\bibitem{liu2022contrastive}
A.~Liu, C.~Zhao, Z.~Yu, J.~Wan, A.~Su, X.~Liu, Z.~Tan, S.~Escalera, J.~Xing, Y.~Liang \emph{et~al.}, ``Contrastive context-aware learning for 3d high-fidelity mask face presentation attack detection,'' \emph{IEEE Transactions on Information Forensics and Security}, vol.~17, pp. 2497--2507, 2022.

\bibitem{george2019biometric}
A.~George, Z.~Mostaani, D.~Geissenbuhler, O.~Nikisins, A.~Anjos, and S.~Marcel, ``Biometric face presentation attack detection with multi-channel convolutional neural network,'' \emph{TIFS}, 2019.

\bibitem{Liu2018Learning}
Y.~Liu, A.~Jourabloo, and X.~Liu, ``Learning deep models for face anti-spoofing: Binary or auxiliary supervision,'' in \emph{CVPR}, 2018.

\bibitem{george2019deep}
A.~George and S.~Marcel, ``Deep pixel-wise binary supervision for face presentation attack detection,'' in \emph{ICB}, 2019.

\bibitem{zhang2020face}
K.-Y. Zhang, T.~Yao, J.~Zhang, Y.~Tai, S.~Ding, J.~Li, F.~Huang, H.~Song, and L.~Ma, ``Face anti-spoofing via disentangled representation learning,'' in \emph{ECCV}, 2020.

\bibitem{yu2020nasfas}
Z.~Yu, J.~Wan, Y.~Qin, X.~Li, S.~Z. Li, and G.~Zhao, ``Nas-fas: Static-dynamic central difference network search for face anti-spoofing,'' in \emph{TPAMI}, 2020.

\bibitem{cai2020drl}
R.~Cai, H.~Li, S.~Wang, C.~Chen, and A.~C. Kot, ``Drl-fas: A novel framework based on deep reinforcement learning for face anti-spoofing,'' \emph{TIFS}, vol.~16, pp. 937--951, 2020.

\bibitem{liu2022disentangling}
A.~Liu, J.~Wan, N.~Jiang, H.~Wang, and Y.~Liang, ``Disentangling facial pose and appearance information for face anti-spoofing,'' in \emph{2022 26th International Conference on Pattern Recognition (ICPR)}.\hskip 1em plus 0.5em minus 0.4em\relax IEEE, 2022, pp. 4537--4543.

\bibitem{Wang_2022_CVPR}
C.-Y. Wang, Y.-D. Lu, S.-T. Yang, and S.-H. Lai, ``Patchnet: A simple face anti-spoofing framework via fine-grained patch recognition,'' pp. 20\,281--20\,290, June 2022.

\bibitem{liu2023towards}
Y.~Liu, Y.~Chen, M.~Gou, C.-T. Huang, Y.~Wang, W.~Dai, and H.~Xiong, ``Towards unsupervised domain generalization for face anti-spoofing,'' in \emph{Proceedings of the IEEE/CVF International Conference on Computer Vision}, 2023.

\bibitem{sun2023rethinking}
Y.~Sun, Y.~Liu, X.~Liu, Y.~Li, and W.-S. Chu, ``Rethinking domain generalization for face anti-spoofing: Separability and alignment,'' in \emph{Proceedings of the IEEE/CVF Conference on Computer Vision and Pattern Recognition}, 2023, pp. 24\,563--24\,574.

\bibitem{zhou2023instance}
Q.~Zhou, K.-Y. Zhang, T.~Yao, X.~Lu, R.~Yi, S.~Ding, and L.~Ma, ``Instance-aware domain generalization for face anti-spoofing,'' in \emph{Proceedings of the IEEE/CVF Conference on Computer Vision and Pattern Recognition}, 2023, pp. 20\,453--20\,463.

\bibitem{srivatsan2023flip}
K.~Srivatsan, M.~Naseer, and K.~Nandakumar, ``Flip: Cross-domain face anti-spoofing with language guidance,'' in \emph{Proceedings of the IEEE/CVF International Conference on Computer Vision}, 2023, pp. 19\,685--19\,696.

\bibitem{le2024grad}
B.~M. Le and S.~S. Woo, ``Gradient alignment for cross-domain face anti-spoofing,'' in \emph{CVPR}, 2024.

\bibitem{zhou2024test}
Q.~Zhou, K.-Y. Zhang, T.~Yao, X.~Lu, S.~Ding, and L.~Ma, ``Test-time domain generalization for face anti-spoofing,'' in \emph{Proceedings of the IEEE/CVF Conference on Computer Vision and Pattern Recognition}, 2024, pp. 175--187.

\bibitem{wang2021safa}
Q.~Wang, L.~Zhang, and B.~Li, ``Safa: Structure aware face animation,'' in \emph{2021 International Conference on 3D Vision (3DV)}.\hskip 1em plus 0.5em minus 0.4em\relax IEEE, 2021, pp. 679--688.

\bibitem{choi2018stargan}
Y.~Choi, M.~Choi, M.~Kim, J.-W. Ha, S.~Kim, and J.~Choo, ``Stargan: Unified generative adversarial networks for multi-domain image-to-image translation,'' in \emph{Proceedings of the IEEE conference on computer vision and pattern recognition}, 2018, pp. 8789--8797.

\bibitem{rosberg2023facedancer}
F.~Rosberg, E.~E. Aksoy, F.~Alonso-Fernandez, and C.~Englund, ``Facedancer: Pose-and occlusion-aware high fidelity face swapping,'' in \emph{Proceedings of the IEEE/CVF winter conference on applications of computer vision}, 2023, pp. 3454--3463.

\bibitem{chen2020simswap}
R.~Chen, X.~Chen, B.~Ni, and Y.~Ge, ``Simswap: An efficient framework for high fidelity face swapping,'' in \emph{Proceedings of the 28th ACM international conference on multimedia}, 2020, pp. 2003--2011.

\bibitem{hong2022depth}
F.-T. Hong, L.~Zhang, L.~Shen, and D.~Xu, ``Depth-aware generative adversarial network for talking head video generation,'' in \emph{Proceedings of the IEEE/CVF conference on computer vision and pattern recognition}, 2022, pp. 3397--3406.

\bibitem{wang2021one}
T.-C. Wang, A.~Mallya, and M.-Y. Liu, ``One-shot free-view neural talking-head synthesis for video conferencing,'' in \emph{Proceedings of the IEEE/CVF conference on computer vision and pattern recognition}, 2021, pp. 10\,039--10\,049.

\bibitem{zou2022making}
J.~Zou, Y.~Duan, B.~Li, W.~Zhang, Y.~Pan, and Z.~Pan, ``Making adversarial examples more transferable and indistinguishable,'' in \emph{Proceedings of the AAAI conference on artificial intelligence}, vol.~36, no.~3, 2022, pp. 3662--3670.

\bibitem{yan2022ila}
C.~W. Yan, T.-H. Cheung, and D.-Y. Yeung, ``Ila-da: Improving transferability of intermediate level attack with data augmentation,'' in \emph{The Eleventh International Conference on Learning Representations}, 2022.

\bibitem{lin2024boosting}
Q.~Lin, C.~Luo, Z.~Niu, X.~He, W.~Xie, Y.~Hou, L.~Shen, and S.~Song, ``Boosting adversarial transferability across model genus by deformation-constrained warping,'' in \emph{Proceedings of the AAAI Conference on Artificial Intelligence}, vol.~38, no.~4, 2024, pp. 3459--3467.

\bibitem{schwinn2023exploring}
L.~Schwinn, R.~Raab, A.~Nguyen, D.~Zanca, and B.~Eskofier, ``Exploring misclassifications of robust neural networks to enhance adversarial attacks,'' \emph{Applied Intelligence}, vol.~53, no.~17, pp. 19\,843--19\,859, 2023.

\bibitem{luo2022frequency}
C.~Luo, Q.~Lin, W.~Xie, B.~Wu, J.~Xie, and L.~Shen, ``Frequency-driven imperceptible adversarial attack on semantic similarity,'' in \emph{Proceedings of the IEEE/CVF conference on computer vision and pattern recognition}, 2022, pp. 15\,315--15\,324.

\bibitem{wei2022towards}
Z.~Wei, J.~Chen, M.~Goldblum, Z.~Wu, T.~Goldstein, and Y.-G. Jiang, ``Towards transferable adversarial attacks on vision transformers,'' in \emph{Proceedings of the AAAI Conference on Artificial Intelligence}, vol.~36, no.~3, 2022, pp. 2668--2676.

\bibitem{zhang2023transferable}
J.~Zhang, Y.~Huang, W.~Wu, and M.~R. Lyu, ``Transferable adversarial attacks on vision transformers with token gradient regularization,'' in \emph{Proceedings of the IEEE/CVF Conference on Computer Vision and Pattern Recognition}, 2023, pp. 16\,415--16\,424.

\bibitem{huang2024consistentid}
J.~Huang, X.~Dong, W.~Song, H.~Li, J.~Zhou, Y.~Cheng, S.~Liao, L.~Chen, Y.~Yan, S.~Liao \emph{et~al.}, ``Consistentid: Portrait generation with multimodal fine-grained identity preserving,'' \emph{arXiv preprint arXiv:2404.16771}, 2024.

\bibitem{guo2024pulid}
Z.~Guo, Y.~Wu, Z.~Chen, L.~Chen, P.~Zhang, and Q.~He, ``Pulid: Pure and lightning id customization via contrastive alignment,'' \emph{arXiv preprint arXiv:2404.16022}, 2024.

\bibitem{wang2024instantid}
Q.~Wang, X.~Bai, H.~Wang, Z.~Qin, A.~Chen, H.~Li, X.~Tang, and Y.~Hu, ``Instantid: Zero-shot identity-preserving generation in seconds,'' \emph{arXiv preprint arXiv:2401.07519}, 2024.

\bibitem{tian2024real}
J.~Tian, C.~Yu, X.~Wang, P.~Chen, Z.~Xiao, J.~Dai, J.~Han, and Y.~Chai, ``Real appearance modeling for more general deepfake detection,'' in \emph{European Conference on Computer Vision}.\hskip 1em plus 0.5em minus 0.4em\relax Springer, 2024, pp. 402--419.

\bibitem{dong2023implicit}
S.~Dong, J.~Wang, R.~Ji, J.~Liang, H.~Fan, and Z.~Ge, ``Implicit identity leakage: The stumbling block to improving deepfake detection generalization,'' in \emph{Proceedings of the IEEE/CVF Conference on Computer Vision and Pattern Recognition}, 2023, pp. 3994--4004.

\bibitem{choi2024exploiting}
J.~Choi, T.~Kim, Y.~Jeong, S.~Baek, and J.~Choi, ``Exploiting style latent flows for generalizing deepfake video detection,'' in \emph{Proceedings of the IEEE/CVF conference on computer vision and pattern recognition}, 2024, pp. 1133--1143.

\bibitem{deb2023unified}
D.~Deb, X.~Liu, and A.~K. Jain, ``Unified detection of digital and physical face attacks,'' in \emph{2023 IEEE 17th International Conference on Automatic Face and Gesture Recognition (FG)}.\hskip 1em plus 0.5em minus 0.4em\relax IEEE, 2023, pp. 1--8.

\bibitem{yu2024benchmarking}
Z.~Yu, R.~Cai, Z.~Li, W.~Yang, J.~Shi, and A.~C. Kot, ``Benchmarking joint face spoofing and forgery detection with visual and physiological cues,'' \emph{IEEE Transactions on Dependable and Secure Computing}, 2024.

\bibitem{fang2024unified}
H.~Fang, A.~Liu, H.~Yuan, J.~Zheng, D.~Zeng, Y.~Liu, J.~Deng, S.~Escalera, X.~Liu, J.~Wan \emph{et~al.}, ``Unified physical-digital face attack detection,'' \emph{arXiv preprint arXiv:2401.17699}, 2024.

\bibitem{Zhang2012A}
Z.~Zhang, J.~Yan, S.~Liu, Z.~Lei, D.~Yi, and S.~Z. Li, ``A face antispoofing database with diverse attacks,'' in \emph{ICB}, 2012.

\bibitem{wen2015face}
D.~Wen, H.~Han, and A.~K. Jain, ``Face spoof detection with image distortion analysis,'' \emph{IEEE TIFS}, 2015.

\bibitem{liu20163d}
S.~Liu, B.~Yang, P.~C. Yuen, and G.~Zhao, ``A 3d mask face anti-spoofing database with real world variations,'' in \emph{Proceedings of the IEEE conference on computer vision and pattern recognition workshops}, 2016, pp. 100--106.

\bibitem{Boulkenafet2017OULU}
Z.~Boulkenafet, J.~Komulainen, L.~Li, X.~Feng, and A.~Hadid, ``Oulu-npu: A mobile face presentation attack database with real-world variations,'' in \emph{FGR}, 2017, pp. 612--618.

\bibitem{li2018unsupervised}
H.~Li, W.~Li, H.~Cao, S.~Wang, F.~Huang, and A.~C. Kot, ``Unsupervised domain adaptation for face anti-spoofing,'' \emph{TIFS}, vol.~13, no.~7, pp. 1794--1809, 2018.

\bibitem{CelebA-Spoof}
Y.~Zhang, Z.~Yin, Y.~Li, G.~Yin, J.~Yan, J.~Shao, and Z.~Liu, ``Celeba-spoof: Large-scale face anti-spoofing dataset with rich annotations,'' in \emph{ECCV}, 2020.

\bibitem{yu2020fas}
Z.~Yu, J.~Wan, Y.~Qin, X.~Li, S.~Z. Li, and G.~Zhao, ``Nas-fas: Static-dynamic central difference network search for face anti-spoofing,'' \emph{IEEE transactions on pattern analysis and machine intelligence}, vol.~43, no.~9, pp. 3005--3023, 2020.

\bibitem{guo2022multi}
X.~Guo, Y.~Liu, A.~Jain, and X.~Liu, ``Multi-domain learning for updating face anti-spoofing models,'' in \emph{ECCV}.\hskip 1em plus 0.5em minus 0.4em\relax Springer, 2022, pp. 230--249.

\bibitem{rostami2021detection}
M.~Rostami, L.~Spinoulas, M.~Hussein, J.~Mathai, and W.~Abd-Almageed, ``Detection and continual learning of novel face presentation attacks,'' in \emph{Proceedings of the IEEE/CVF international conference on computer vision}, 2021, pp. 14\,851--14\,860.

\bibitem{dufour2019deepfakes}
N.~Dufour, A.~Gully, P.~Karlsson, A.~Vorbyov, T.~Leung, J.~Childs, and C.~Bregler, ``Deepfakes detection dataset,'' \emph{Google and Jigsaw}, 2019.

\bibitem{dolhansky2019deepfake}
B.~Dolhansky, R.~Howes, B.~Pflaum, N.~Baram, and C.~C. Ferrer, ``The deepfake detection challenge (dfdc) preview dataset,'' \emph{arXiv preprint arXiv:1910.08854}, 2019.

\bibitem{rossler2019faceforensics++}
A.~Rossler, D.~Cozzolino, L.~Verdoliva, C.~Riess, J.~Thies, and M.~Nie{\ss}ner, ``Faceforensics++: Learning to detect manipulated facial images,'' in \emph{Proceedings of the IEEE/CVF international conference on computer vision}, 2019, pp. 1--11.

\bibitem{li2020celeb}
Y.~Li, X.~Yang, P.~Sun, H.~Qi, and S.~Lyu, ``Celeb-df: A large-scale challenging dataset for deepfake forensics,'' in \emph{Proceedings of the IEEE/CVF conference on computer vision and pattern recognition}, 2020, pp. 3207--3216.

\bibitem{Jiang_2020_CVPR}
L.~Jiang, R.~Li, W.~Wu, C.~Qian, and C.~C. Loy, ``Deeperforensics-1.0: A large-scale dataset for real-world face forgery detection,'' in \emph{Proceedings of the IEEE/CVF Conference on Computer Vision and Pattern Recognition (CVPR)}, June 2020.

\bibitem{Zhou_2021_CVPR}
T.~Zhou, W.~Wang, Z.~Liang, and J.~Shen, ``Face forensics in the wild,'' in \emph{Proceedings of the IEEE/CVF Conference on Computer Vision and Pattern Recognition (CVPR)}, June 2021, pp. 5778--5788.

\bibitem{he2021forgerynet}
Y.~He, Y.~Zhang, Y.~Liu, W.~Wang, Y.~Quan, S.~Lyu, and Z.~Liu, ``Forgerynet: A versatile benchmark for face forgery detection and localization,'' in \emph{Proceedings of the IEEE/CVF International Conference on Computer Vision (ICCV)}, 2021, pp. 15\,042--15\,052.

\bibitem{liu2021casia}
A.~Liu, Z.~Tan, J.~Wan, S.~Escalera, G.~Guo, and S.~Z. Li, ``Casia-surf cefa: A benchmark for multi-modal cross-ethnicity face anti-spoofing,'' in \emph{WACV}, 2021, pp. 1179--1187.

\bibitem{zou2024softmoe}
H.~Zou, C.~Du, H.~Zhang, Y.~Zhang, A.~Liu, J.~Wan, and Z.~Lei, ``La-softmoe clip for unified physical-digital face attack detection,'' in \emph{2024 IEEE International Joint Conference on Biometrics (IJCB)}.\hskip 1em plus 0.5em minus 0.4em\relax IEEE, 2024, pp. 1--11.

\bibitem{chen2025mixture}
S.~Chen, A.~Liu, J.~Zheng, J.~Wan, K.~Peng, S.~Escalera, and Z.~Lei, ``Mixture-of-attack-experts with class regularization for unified physical-digital face attack detection,'' in \emph{Proceedings of the AAAI Conference on Artificial Intelligence}, vol.~39, no.~2, 2025, pp. 2195--2203.

\bibitem{li2025fa}
Y.~Li, N.~Li, A.~Liu, H.~Ma, L.~Yang, X.~Chen, Z.~Liang, Y.~Liang, J.~Wan, and Z.~Lei, ``Fa\^{}$\{$3$\}$-clip: Frequency-aware cues fusion and attack-agnostic prompt learning for unified face attack detection,'' \emph{arXiv preprint arXiv:2504.00454}, 2025.

\bibitem{timesler2020facenet_pytorch}
\BIBentryALTinterwordspacing
Timesler, ``facenet-pytorch,'' 2020. [Online]. Available: \url{https://github.com/timesler/facenet-pytorch}
\BIBentrySTDinterwordspacing

\bibitem{9851423ghost}
A.~Groshev, A.~Maltseva, D.~Chesakov, A.~Kuznetsov, and D.~Dimitrov, ``Ghost—a new face swap approach for image and video domains,'' \emph{IEEE Access}, vol.~10, pp. 83\,452--83\,462, 2022.

\bibitem{ren2023pbidrinsightface}
X.~Ren, A.~Lattas, B.~Gecer, J.~Deng, C.~Ma, and X.~Yang, ``Facial geometric detail recovery via implicit representation,'' in \emph{2023 IEEE 17th International Conference on Automatic Face and Gesture Recognition (FG)}, 2023.

\bibitem{shiohara2023blendface}
K.~Shiohara, X.~Yang, and T.~Taketomi, ``Blendface: Re-designing identity encoders for face-swapping,'' in \emph{Proceedings of the IEEE/CVF International Conference on Computer Vision}, 2023, pp. 7634--7644.

\bibitem{baliah2024realistic}
S.~Baliah, Q.~Lin, S.~Liao, X.~Liang, and M.~H. Khan, ``Realistic and efficient face swapping: A unified approach with diffusion models,'' \emph{arXiv preprint arXiv:2409.07269}, 2024.

\bibitem{siarohin2021motion}
A.~Siarohin, O.~J. Woodford, J.~Ren, M.~Chai, and S.~Tulyakov, ``Motion representations for articulated animation,'' in \emph{Proceedings of the IEEE/CVF Conference on Computer Vision and Pattern Recognition}, 2021, pp. 13\,653--13\,662.

\bibitem{gao2021information}
G.~Gao, H.~Huang, C.~Fu, Z.~Li, and R.~He, ``Information bottleneck disentanglement for identity swapping,'' in \emph{Proceedings of the IEEE/CVF conference on computer vision and pattern recognition}, 2021, pp. 3404--3413.

\bibitem{zhu2021one}
Y.~Zhu, Q.~Li, J.~Wang, C.-Z. Xu, and Z.~Sun, ``One shot face swapping on megapixels,'' in \emph{Proceedings of the IEEE/CVF conference on computer vision and pattern recognition}, 2021, pp. 4834--4844.

\bibitem{wang2021hififace}
Y.~Wang, X.~Chen, J.~Zhu, W.~Chu, Y.~Tai, C.~Wang, J.~Li, Y.~Wu, F.~Huang, and R.~Ji, ``Hififace: 3d shape and semantic prior guided high fidelity face swapping,'' \emph{arXiv preprint arXiv:2106.09965}, 2021.

\bibitem{zhao2023diffswap}
W.~Zhao, Y.~Rao, W.~Shi, Z.~Liu, J.~Zhou, and J.~Lu, ``Diffswap: High-fidelity and controllable face swapping via 3d-aware masked diffusion,'' in \emph{Proceedings of the IEEE/CVF Conference on Computer Vision and Pattern Recognition}, 2023, pp. 8568--8577.

\bibitem{deeplivecam}
\BIBentryALTinterwordspacing
hacksider. (2024) Deep-live-cam: Real-time face swapping and animation tool. Accessed: 2025-05-18. [Online]. Available: \url{https://github.com/hacksider/Deep-Live-Cam}
\BIBentrySTDinterwordspacing

\bibitem{roop2023}
{s0md3v}, ``Roop: One-image face swap on video,'' \url{https://github.com/s0md3v/roop}, 2023, accessed: 2025-05-18.

\bibitem{rony2021augmented}
J.~Rony, E.~Granger, M.~Pedersoli, and I.~Ben~Ayed, ``Augmented lagrangian adversarial attacks,'' in \emph{Proceedings of the IEEE/CVF International Conference on Computer Vision}, 2021, pp. 7738--7747.

\bibitem{duan2021advdrop}
R.~Duan, Y.~Chen, D.~Niu, Y.~Yang, A.~K. Qin, and Y.~He, ``Advdrop: Adversarial attack to dnns by dropping information,'' in \emph{Proceedings of the IEEE/CVF International Conference on Computer Vision}, 2021, pp. 7506--7515.

\bibitem{croce2020reliable}
F.~Croce and M.~Hein, ``Reliable evaluation of adversarial robustness with an ensemble of diverse parameter-free attacks,'' in \emph{International conference on machine learning}.\hskip 1em plus 0.5em minus 0.4em\relax PMLR, 2020, pp. 2206--2216.

\bibitem{madry2017towards}
A.~Madry, ``Towards deep learning models resistant to adversarial attacks,'' \emph{arXiv preprint arXiv:1706.06083}, 2017.

\bibitem{lin2019nesterov}
J.~Lin, C.~Song, K.~He, L.~Wang, and J.~E. Hopcroft, ``Nesterov accelerated gradient and scale invariance for adversarial attacks,'' \emph{arXiv preprint arXiv:1908.06281}, 2019.

\bibitem{gao2020patch}
L.~Gao, Q.~Zhang, J.~Song, and H.~T. Shen, ``Patch-wise++ perturbation for adversarial targeted attacks,'' \emph{arXiv preprint arXiv:2012.15503}, 2020.

\bibitem{pomponi2022pixle}
J.~Pomponi, S.~Scardapane, and A.~Uncini, ``Pixle: a fast and effective black-box attack based on rearranging pixels,'' in \emph{2022 International Joint Conference on Neural Networks (IJCNN)}.\hskip 1em plus 0.5em minus 0.4em\relax IEEE, 2022, pp. 1--7.

\bibitem{wang2021enhancing}
X.~Wang and K.~He, ``Enhancing the transferability of adversarial attacks through variance tuning,'' in \emph{Proceedings of the IEEE/CVF conference on computer vision and pattern recognition}, 2021, pp. 1924--1933.

\bibitem{wang2021demiguise}
Y.~Wang, S.~Wu, W.~Jiang, S.~Hao, Y.-a. Tan, and Q.~Zhang, ``Demiguise attack: Crafting invisible semantic adversarial perturbations with perceptual similarity,'' \emph{arXiv preprint arXiv:2107.01396}, 2021.

\bibitem{li2024photomaker}
Z.~Li, M.~Cao, X.~Wang, Z.~Qi, M.-M. Cheng, and Y.~Shan, ``Photomaker: Customizing realistic human photos via stacked id embedding,'' in \emph{Proceedings of the IEEE/CVF Conference on Computer Vision and Pattern Recognition}, 2024, pp. 8640--8650.

\bibitem{ye2023ip}
H.~Ye, J.~Zhang, S.~Liu, X.~Han, and W.~Yang, ``Ip-adapter: Text compatible image prompt adapter for text-to-image diffusion models,'' \emph{arXiv preprint arXiv:2308.06721}, 2023.

\bibitem{radford2021learning}
A.~Radford, J.~W. Kim, C.~Hallacy, A.~Ramesh, G.~Goh, S.~Agarwal, G.~Sastry, A.~Askell, P.~Mishkin, J.~Clark \emph{et~al.}, ``Learning transferable visual models from natural language supervision,'' in \emph{International conference on machine learning}.\hskip 1em plus 0.5em minus 0.4em\relax PMLR, 2021, pp. 8748--8763.

\bibitem{vaswani2017attention}
A.~Vaswani, ``Attention is all you need,'' \emph{Advances in Neural Information Processing Systems}, 2017.

\bibitem{dosovitskiy2020vit}
A.~Dosovitskiy, L.~Beyer, A.~Kolesnikov, D.~Weissenborn, X.~Zhai, T.~Unterthiner, M.~Dehghani, M.~Minderer, G.~Heigold, S.~Gelly, J.~Uszkoreit, and N.~Houlsby, ``An image is worth 16x16 words: Transformers for image recognition at scale,'' \emph{ICLR}, 2021.

\bibitem{zhou2022learning}
K.~Zhou, J.~Yang, C.~C. Loy, and Z.~Liu, ``Learning to prompt for vision-language models,'' \emph{International Journal of Computer Vision}, vol. 130, no.~9, pp. 2337--2348, 2022.

\bibitem{jia2022visual}
M.~Jia, L.~Tang, B.-C. Chen, C.~Cardie, S.~Belongie, B.~Hariharan, and S.-N. Lim, ``Visual prompt tuning,'' in \emph{European Conference on Computer Vision}.\hskip 1em plus 0.5em minus 0.4em\relax Springer, 2022, pp. 709--727.

\bibitem{yu2024visual}
Z.~Yu, R.~Cai, Y.~Cui, A.~Liu, and C.~Chen, ``Visual prompt flexible-modal face anti-spoofing,'' \emph{IEEE Transactions on Dependable and Secure Computing}, 2024.

\bibitem{yu2020searching}
Z.~Yu, C.~Zhao, Z.~Wang, Y.~Qin, Z.~Su, X.~Li, F.~Zhou, and G.~Zhao, ``Searching central difference convolutional networks for face anti-spoofing,'' in \emph{Proceedings of the IEEE/CVF conference on computer vision and pattern recognition}, 2020, pp. 5295--5305.

\bibitem{liu2024cfpl}
A.~Liu, S.~Xue, J.~Gan, J.~Wan, Y.~Liang, J.~Deng, S.~Escalera, and Z.~Lei, ``Cfpl-fas: Class free prompt learning for generalizable face anti-spoofing,'' in \emph{Proceedings of the IEEE/CVF Conference on Computer Vision and Pattern Recognition}, 2024, pp. 222--232.

\bibitem{kong2024moe}
C.~Kong, A.~Luo, P.~Bao, Y.~Yu, H.~Li, Z.~Zheng, S.~Wang, and A.~C. Kot, ``Moe-ffd: Mixture of experts for generalized and parameter-efficient face forgery detection,'' \emph{arXiv preprint arXiv:2404.08452}, 2024.

\bibitem{nguyen2024laa}
D.~Nguyen, N.~Mejri, I.~P. Singh, P.~Kuleshova, M.~Astrid, A.~Kacem, E.~Ghorbel, and D.~Aouada, ``Laa-net: Localized artifact attention network for quality-agnostic and generalizable deepfake detection,'' in \emph{Proceedings of the IEEE/CVF Conference on Computer Vision and Pattern Recognition}, 2024, pp. 17\,395--17\,405.

\bibitem{afchar2018mesonet}
D.~Afchar, V.~Nozick, J.~Yamagishi, and I.~Echizen, ``Mesonet: a compact facial video forgery detection network,'' in \emph{WIFS}, 2018.

\bibitem{chollet2017xception}
F.~Chollet, ``Xception: Deep learning with depthwise separable convolutions,'' in \emph{Proceedings of the IEEE Conference on Computer Vision and Pattern Recognition (CVPR)}, 2017, pp. 1251--1258.

\end{thebibliography}
%




\end{document}